\documentclass[10pt,journal,compsoc]{IEEEtran}
\ifCLASSOPTIONcompsoc
  \usepackage[nocompress]{cite}
\else
  \usepackage{cite}
\fi
\ifCLASSINFOpdf
\else
\fi

\usepackage{threeparttable}

\usepackage{amsmath,amsfonts}
\usepackage{algorithmic}
\newtheorem{MyDef}{Definition}[subsection]
\newtheorem{MyTheory}{Theorem}
\usepackage{multirow}
\usepackage{algorithmic}
\usepackage{algorithm}
\usepackage{array}
\usepackage{colortbl} 
\usepackage{xcolor} 
\usepackage[caption=false,font=normalsize,labelfont=sf,textfont=sf]{subfig}
\usepackage{textcomp}
\usepackage{stfloats}
\usepackage{url}
\usepackage{verbatim}
\usepackage{graphicx}
\usepackage{cite}

\usepackage[top=0.4in,bottom=0.4in,left=0.4in,textwidth=7.7in] {geometry}
\usepackage{xspace}
\makeatletter 
\DeclareRobustCommand\onedot{\futurelet\@let@token\@onedot}
\def\@onedot{\ifx\@let@token.\else.\null\fi\xspace}

\def\eg{\emph{e.g}\onedot} 
\def\ie{\emph{i.e}\onedot} 
 
\def\etc{\emph{etc}\onedot} 
 
\def\etal{\emph{et al}\onedot}
\makeatother

\usepackage[pagebackref=false,breaklinks=true,linkcolor=red,anchorcolor=blue, citecolor=green,letterpaper=true,colorlinks,bookmarks=false]{hyperref}
\begin{document}


\title{Step-wise Distribution-aligned Style Prompt Tuning for  Source-Free Cross-domain Few-shot Learning}

\author{Huali Xu, Li Liu, Tianpeng Liu, Shuaifeng Zhi, Shuzhou Sun, Ming-Ming Cheng 
\thanks{This work was partially supported by National Key Research and Development Program of China No. 2021YFB3100800, the Academy of Finland under grant Infotech Project FRAGES,
the National Natural Science Foundation of China under Grant 62376283, 62531026, 62201603, 62571535, the HNNSF under Grant 2025JJ40060 and the CPSF under Grant 2023TQ0088, GZC20233539.}
\thanks{Huali Xu is with the College of Computer Science, Nankai University, Tianjin 300071, China, and also with the Center for Machine Vision and
Signal Analysis (CMVS), University of Oulu, 90570 Oulu, Finland (email: huali.xu@oulu.fi).}
\thanks{Shuzhou Sun is with the Center for Machine Vision and Signal Analysis (CMVS), University of Oulu, Oulu 90570, Finland (email: shuzhou.sun@oulu.fi).}
\thanks{Shuaifeng Zhi, Tianpeng Liu, and Li Liu are with the College of Electronic Science and Technology, National University of Defense Technology, Changsha, 410073, China (email: zhishuaifeng@outlook.com; lilyliu\_nudt@163.com).}
\thanks{Corresponding authors: Li Liu.}
\thanks{Ming-Ming Cheng (cmm@nankai.edu.cn) is with the College of Computer Science, Nankai University, Tianjin, TKLNDST, China.}}

\markboth{Submit to IEEE Transactions on Pattern Analysis and Machine Intelligence}%
{Xu \MakeLowercase{\textit{et al.}}: A Sample Article Using IEEEtran.cls for IEEE Journals}

\IEEEpubid{0000--0000/00\$00.00~\copyright~2021 IEEE}

\IEEEtitleabstractindextext{%
\begin{abstract}
Existing cross-domain few-shot learning (CDFSL) methods, which develop training strategies in the source domain to enhance model transferability, face challenges when applied to large-scale pre-trained models (LMs), as their source domains and training strategies are not accessible. Besides, fine-tuning LMs specifically for CDFSL requires substantial computational resources, which limits their practicality. Therefore, this paper investigates the source-free CDFSL (SF-CDFSL) problem to solve the few-shot learning (FSL) task in target domain using only a pre-trained model and a few target samples, without requiring source data or training strategies. However, the inaccessibility of source data prevents explicitly reducing the domain gaps between the source and target. To tackle this challenge, this paper proposes a novel approach, \textbf{Step}-wise Distribution-aligned \textbf{S}tyle \textbf{P}rompt \textbf{T}uning (\textbf{StepSPT}), to implicitly narrow the domain gaps from the perspective of prediction distribution optimization. StepSPT initially proposes a style prompt that adjusts the target samples to mirror the expected distribution. Furthermore, StepSPT tunes the style prompt and classifier by exploring a dual-phase optimization process (external and internal processes). In the external process, a step-wise distribution alignment strategy is introduced to tune the proposed style prompt by factorizing the prediction distribution optimization problem into the multi-step distribution alignment problem. In the internal process, the classifier is updated via standard cross-entropy loss. Evaluation on 5 datasets illustrates the superiority of StepSPT over existing prompt tuning-based methods and state-of-the-art methods (SOTAs). Furthermore, ablation studies and performance analyzes highlight the efficacy of StepSPT. The code will be made public at \url{https://github.com/xuhuali-mxj/StepSPT}.
\end{abstract}

}

\maketitle

\section{Introduction}
\IEEEPARstart{C}{ross}-domain few-shot learning (CDFSL)~\cite{xu2023deep} aims to address the target task with limited data by leveraging a vast dataset from a different domain, \ie source domain. Existing CDFSL methods~\cite{tseng2020cross,phoo2020self,islam2021dynamic,xu2022cross,fu2023styleadv,zhao2023fs,li2022ranking,zhao2023dual} tackle the target FSL task by utilizing massive data from alternative (source) domains and the corresponding training or adapt strategy, which requires the source data to be accessible and allows the training strategy to be designed. However, these methods may not always be feasible in real-world scenarios~\cite{liang2020we} due to the following reasons: (1) Concerns regarding confidentiality, privacy, and copyright may render the source dataset inaccessible. (2) The computational burden of training with a large source dataset is especially challenging for edge devices. (3) Large-scale pre-training models (LMs)~\cite{radford2021learning,dosovitskiy2020image,liu2020feature,wang2023large,devlin2018bert,brown2020language,pmlr-v202-shu23a,11016924}, renowned for robust generalization, are crucial for CDFSL. However, their application is hindered by the inaccessibility of source data and the inability to design training strategies. 

\begin{figure*}[!t]
\centering
\includegraphics[width=\linewidth]{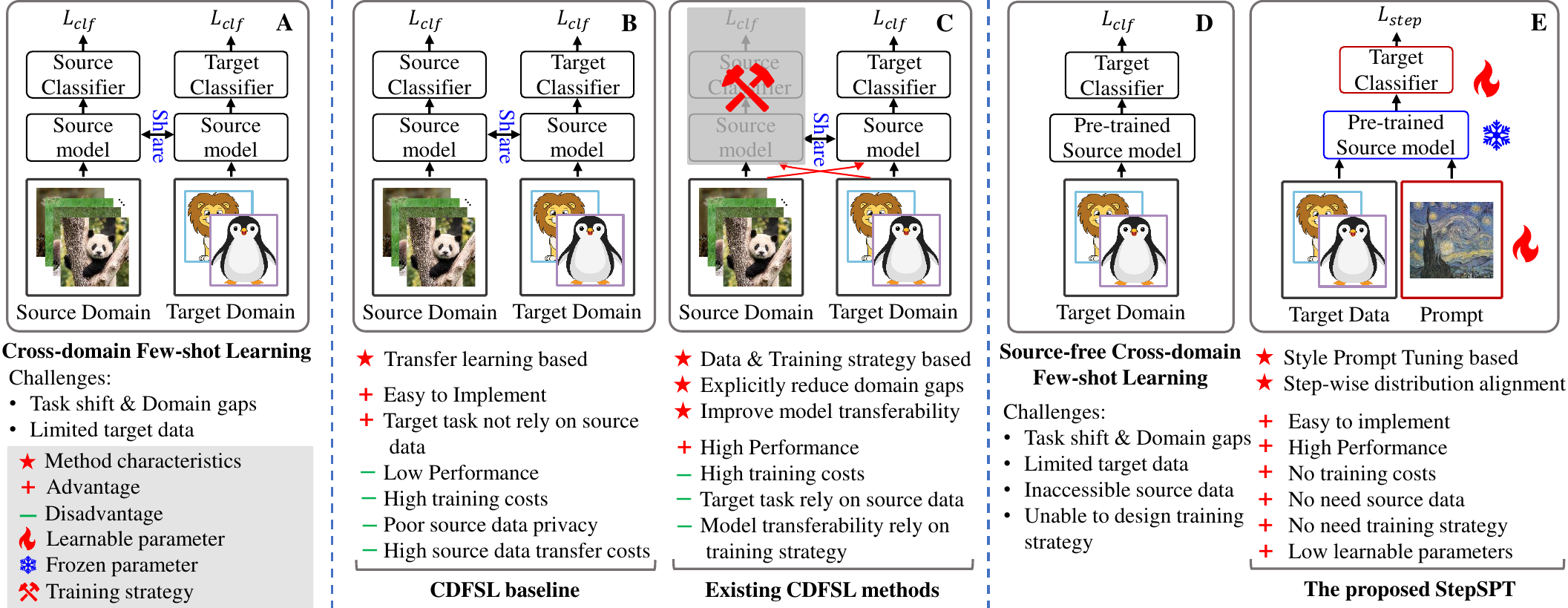}
\vspace{-3mm}
\caption{The difference between CDFSL and SF-CDFSL. \textbf{A:} The CDFSL setup. \textbf{B \& C:} The CDFSL baselines and existing solutions, \eg introducing the source/target data into the adapt/training process, or exploring the training strategy. \textbf{D:} The SF-CDFSL setup. \textbf{E:} The proposed StepSPT makes both the pretrained source model and target domain adapt to each other through the learnable style prompt and classifier.}
\vspace{-5mm}
\label{setup}
\end{figure*}

To address the above mentioned problems, researchers delve into the exploration of a novel Source-Free CDFSL (SF-CDFSL)~\cite{xu2024enhancing} problem. SF-CDFSL addresses the FSL task in the target domain exclusively through an existing pretrained model (referred to as the source model), without access to source data or training strategies. Figure~\ref{setup} depicts the differences between the vanilla CDFSL and SF-CDFSL. By relaxing the requirements on accessing source domain data and designing training strategies, exploration of SF-CDFSL, on the one hand, protects source data privacy, which promotes the development of fields with high data privacy requirements, such as medicine~\cite{wang2017chestx,codella2019skin}, remote sensing~\cite{li2024predicting,zhou2024diffdet4sar}, \etc On the other hand, SF-CDFSL can ignore the source data transmission costs~\cite{liang2020we,li2024comprehensive}. Moreover, the reduced restrictions on training strategy design make it feasible to apply LMs to CDFSL tasks. However, in addition to the challenges inherited from CDFSL, SF-CDFSL has unique challenges~\cite{xu2024enhancing}. Firstly, in contrast to CDFSL, which utilizes both source and target data, SF-CDFSL relies solely on scarce labeled target data to tackle the FSL challenge, prone to overfitting. Secondly, explicitly aligning source and target distributions becomes impossible due to the lack of access to source data.

To address these challenges, this paper aims to optimize the prediction distribution to implicitly reduce the gaps between source and target domains. 
Leveraging the robust generalization capabilities of LMs as the source model, Step-wise Distribution-aligned Style Prompt Tuning (StepSPT) is proposed to solve the FSL task in the target domain. First, inspired by the ability of batch normalization to adjust style and affect data distribution~\cite{yazdanpanah2022visual}, StepSPT designs a learnable style prompt to reduce the difference between the target and expected distributions, while also constraining the model's hypothesis space to prevent overfitting. Second, StepSPT utilizes a dual-phase optimization process consisting of external and internal stages. In both phases, the parameters of the LMs remain frozen, while the style prompt and classifier parameters are updated iteratively, ensuring computational efficiency and preserving the integrity of source model.
In the external phase, the classifier parameters are frozen, and transductive learning is introduced to achieve the distribution alignment. Through in-depth analysis of the target domain distribution optimization problem~\cite{kumar2020understanding,tang2023source,10167762}, we identified that the large domain gap between source and target distributions makes direct optimization challenging. To tackle this, we reformulate the problem as a step-wise distribution alignment strategy, which progressively reduces the domain gap by constraining the difference between adjacent steps. This strategy enables gradual alignment of the target distribution with the ideal distribution, optimizing prediction performance incrementally. The query set is introduced into the external process due to the positive effects of transductive learning. Since the data in the query set are unlabeled, the step-wise distribution alignment strategy uses the data distribution of the previous step as a reference and aligns the distribution of the current step to it. This implies that the accurate guidance provided by the data distribution from the previous step is crucial. To ensure the distribution optimization of the subsequent step is not misled by the previous step's distribution during the alignment process, a credible group is formed in the prior step to identify and select trustworthy features for alignment. The prompt is learned through multiple alignment steps, allowing the target distribution to gradually approach the ideal distribution. In the internal phase, a standard meta-training strategy is applied to the support set to update the classifier while keeping the prompt parameters fixed. This phase helps the model adapt to the specific characteristics of the target domain. The contributions of this paper are as follows:
\begin{itemize}
    \item We focus on a new SF-CDFSL problem, and transform the domain alignment challenge between the source and target domains into a target distribution optimization problem. Furthermore, we provide a theoretical analysis and guidance to address this distribution optimization. 
    \item We propose the Step-wise Distribution-aligned Style Prompt Tuning (StepSPT) method to tackle the SF-CDFSL problem. StepSPT introduces a style prompt to adjust the target distribution and employs a dual-phase optimization process including external and internal stages, where the style prompt and classifier are alternately updated in the external and internal processes. In the external process, we make the target distribution close to the ideal distribution, while let the model adapt to the target FSL task in the internal process. Specifically, in external process, through theoretical analysis, we turn the difficult one-step alignment into simple multi-step alignment, Based on the this, we explore a step-wise distribution alignment strategy to learn the proposed style prompt. Besides, a credible group is introduced to avoid misleading information during alignment, while the traditional cross-entropy loss is used in the internal process to update the classifier. 
    \item Extensive evaluation on 5 datasets indicates the superior performance of the proposed StepSPT, with detailed ablation study illustrating the contribution of each component. 
\end{itemize}

\section{Related Work}
\subsection{Cross-domain Few-shot Learning}
Since Guo \etal~\cite{guo2020broader} set the BSCD-FSL benchmark and Tseng \etal~\cite{tseng2020cross} conceptualized it, many studies have appeared in CDFSL~\cite{sun2021explanation,phoo2020self}. Most of them focused on pre-training strategies in CDFSL~\cite{phoo2020self,islam2021dynamic,xu2022cross,fu2021meta}. A common practice is to introduce additional data during the training phase to facilitate the acquisition of shared knowledge between the source and the target domain~\cite{phoo2020self,islam2021dynamic,fu2022generalized,fu2021meta}. For example,~\cite{phoo2020self} learns the source domain representation with a self-training strategy by using unlabeled target data. Following the above method,~\cite{islam2021dynamic} proposes a dynamic distillation-based approach, utilizing unlabeled images from the target domain. Also,~\cite{fu2021meta} uses a small amount of labeled target data for training. They introduced the meta-FDMixup network, which helps the model by mixing up different data and separating important features. In addition, other CDFSL solutions incorporate additional information, such as style, during the pre-training phase~\cite{xu2022cross,fu2022wave,zhang2022free,fu2023styleadv}.~\cite{xu2022cross} introduces an ISSNet, which enhances model generalization by applying styles from unlabeled data to labeled data. ~\cite{fu2022wave} tackle CDFSL by analyzing style variations, breaking down images into simpler components using wavelet transform to better understand their styles. Furthermore, they also develop a training method called StyleAdv, focusing on meta style adversarial training for CDFSL. Besides,~\cite{zhang2022free} proposes the SET-RCL method, which adapts to new styles by considering both data and model adjustments to mimic the styles of unknown domains. Additionally, ~\cite{graph1} introduces a Gia-CFSL framework, combining few-shot learning with domain alignment in hyperspectral image classification through graph information, to bridge the gap between different domains effectively. 

Currectly, CDFSL technology focus on adaptation strategies in the target domain. This is in addition to in-depth analyses of training strategies in the source domain~\cite{li2022ranking,zhao2023dual}. ~\cite{li2022ranking} use traditional distance-based classifiers and image retrieval views, they apply a reranking procedure to refine the target distance matrix by identifying k-reciprocal neighbors in the task.~\cite{zhao2023dual} proposes a dual adaptive representation alignment approach that leverages prototypical feature alignment and normalized distribution alignment to enable fast adaptation of meta-learners with very few-shot samples, leading to SOTA results on multiple benchmarks in the field of few-shot learning. All of these methods have yielded exceptional results, significantly advancing the field of CDFSL. These novel perspectives emphasize the technical potential of exploring the adapt strategy and not relying on the CDFSL training strategy. It is worth noting that all these algorithms require access to source data. In contrast, our SF-CDFSL prevents source data access, eliminating privacy and transmission issues. Furthermore, SF-CDFSL provides a robust method and framework for the application of LMs in CDFSL tasks.

\subsection{Source-free Domain Adaptation}
Traditional domain adaptation methods~\cite{10506994,9736609} often rely on access to source domain data during adaptation, leading to the emergence of Source-Free Domain Adaptation (SFDA). The initial foray into Source-Free Domain Adaptation (SFDA)\cite{liang2020we} posits that source data is dispensable, advocating for the retention of frozen source classifier parameters to align target representations with source hypotheses' predictions. Subsequently,~\cite{roy2022uncertainty} leverages uncertainty in source model predictions to guide target domain adaptation. Additionally, ~\cite{yang2021exploiting} observes that despite poor performance of the source model in the target domain due to domain gaps, the generated features tend to cluster, prompting the utilization of nearest neighbor methods to minimize distances between similar samples. Some other methods explore the domain-invariant information. For instance, \cite{Wang_2022_CVPR} proposes the Domain-Invariant Parameter Exploration (DIPE) method to identify domain-invariant parameters within the source model, thereby enabling the generation of domain-invariant representations. \cite{Zhang_2023_CVPR} proposes transferring the domain-invariant class relationships to fully leverage the knowledge from the source domain. While some methods solve SFDA though clustering. Specifically, by treating SFDA as an unsupervised clustering problem and leveraging the idea that local neighbors in the feature space are likely to have more similar predictions, \cite{yang2022attracting} proposes optimizing an objective that enforces prediction consistency among neighboring features. \cite{zhou2024source} constructs a set of semantic class prototypes to cluster target features and assign pseudo-labels, facilitating effective distribution alignment by using these pseudo-labeled target samples along with the class prototypes. Moreover, \cite{Tang_2024_CVPR} proposes a Distilling Multimodal Foundation Model (DIFO) approach that maximizes mutual information with the target model using prompt learning, while simultaneously distilling the knowledge model into the target model. Research indicates that SFDA frequently outperforms conventional Unsupervised Domain Adaptation (UDA)~\cite{liang2020we}. However, SFDA typically have identical source and target tasks. In contrast, SF-CDFSL encounters domain gaps and task disparities between the source model and the target task, making it more challenging.

\subsection{Prompt Learning}
Despite the outstanding performance of current large-scale pre-training models (LMs)~\cite{radford2021learning,dosovitskiy2020image,liu2020feature} across a variety of downstream tasks, their massive parameters result in a finetuning process that is both time-consuming and resource-intensive. Therefore, researchers have explored prompt tuning, an efficient approach that fine-tunes a small part of the input prompts instead of all parameters of model. Existing prompt tuning methods~\cite{jia2022visual,zhou2022learning,zhou2022conditional,ge2023domain,zhu2023prompt,xin2024mmap,ma2023prod,wu2024task} typically employ an automated prompt, input alongside data into LMs, to accommodate downstream tasks. Among them,~\cite{jia2022visual,ma2023prod} updates a prompt and classifier simultaneously to adjust to the downstream tasks. Furthermore,~\cite{zhou2022learning,zhou2022conditional} utilize CLIP as a backbone introduce text prompts, expanding the versatility of prompt tuning. However, due to limited knowledge of target domains, LMs often perform poorly in cross-domain tasks, especially for distant domain. Thus, creating an adaptive prompts for different domains is key to improving LMs' performance in cross-domain tasks.

\section{Methods}
\subsection{Overview}
The primary goal of CDFSL is to address the FSL task in the target domain with the help of source task with sufficient data. Specifically, given a source domain $\mathcal{D}^s$ with task $\mathcal{T}^s$, and a target domain $\mathcal{D}^t$ with a few-shot learning (FSL) task $\mathcal{T}^t$ ($\mathcal{D}^s \neq \mathcal{D}^t$, $\mathcal{T}^s \cap \mathcal{T}^t = \emptyset$), CDFSL first leverages labeled data from $\mathcal{D}^s$ for solving $\mathcal{T}^s$. The target domain $\mathcal{D}^t$ includes a support set $S=\left\{ x_{i}, y_{i} \right\}^{N \times K}_{i=1}$ and a query set $Q=\left\{ x_{i} \right\}^{m-(N \times K)}_{i=1}$, \textit{m} means the sample numbers in $\mathcal{D}^t$. Here, $x_{i}$ and $y_{i}$ denote the training samples and their labels in $S$, forming an $N$-way $K$-shot FSL task. The query set $Q$ contains samples from the same $N$ classes as $S$, used to evaluate the FSL model. The goal of CDFSL is to learn a predictive function $f$ for $\mathcal{T}^t$ using both the limited samples in $\mathcal{D}^t$ and the knowledge from $(\mathcal{D}^s, \mathcal{T}^s)$.
The main difference between SF-CDFSL and CDFSL is the accessibility of $\mathcal{D}^s$. In SF-CDFSL, the source data is inaccessible, and the aim is to address the target FSL task $\mathcal{T}^t$ using only a few labeled data in $\mathcal{D}^t$ and a pretrained source model $\theta$.
Existing methods~\cite{fu2022wave,fu2023styleadv,xu2023deep,zhang2022free} focus on explicitly reducing the distance between the source and target domains. However, these approaches face two key challenges in SF-CDFSL: (1) SF-CDFSL relies only on limited labeled target data to tackle the FSL task, and (2) explicitly aligning source and target distributions to bridge domain disparities becomes infeasible due to the unknown source data distribution.

\begin{figure*}[!t]
\centering
\includegraphics[width=0.8\linewidth]{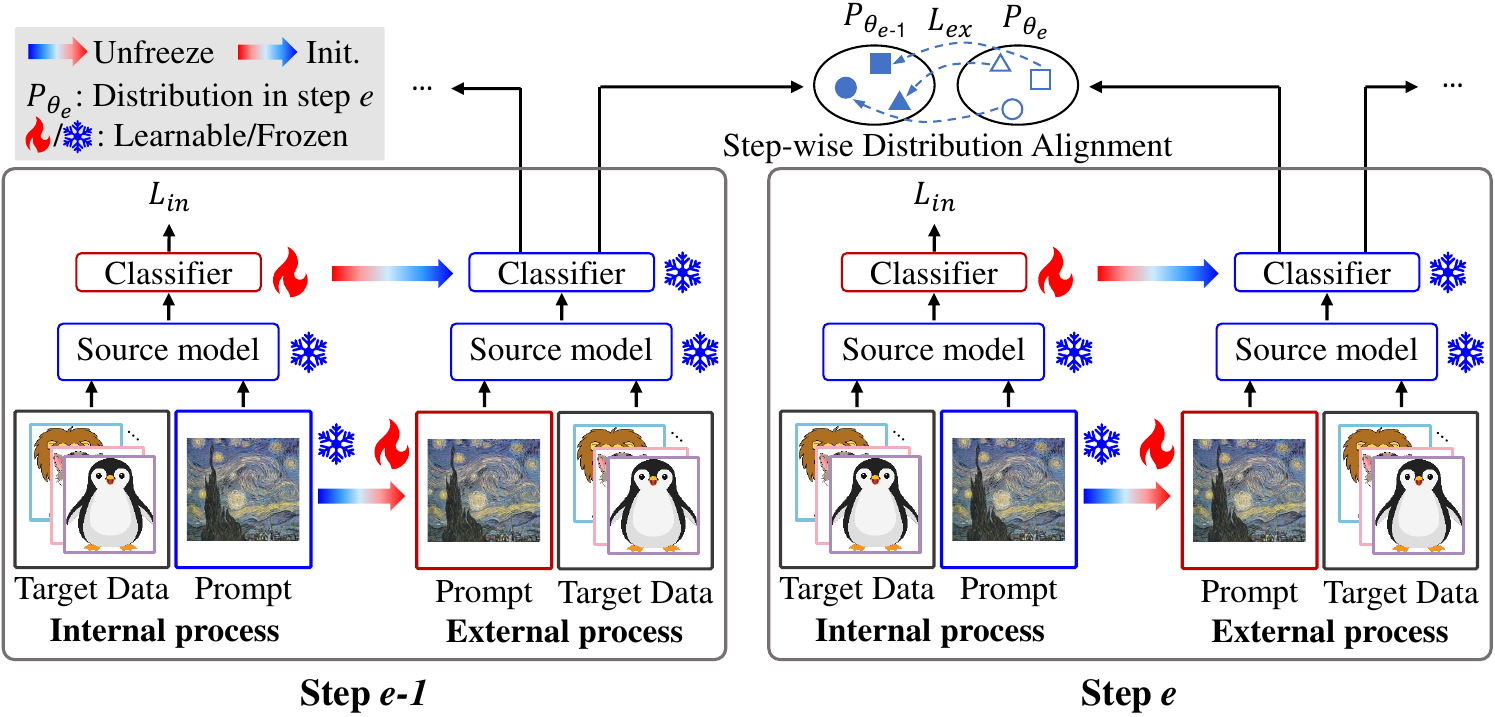}
\vspace{-3mm}
\caption{Overview of StepSPT. In the internal process, only the classifier is updated. In the external process, the proposed style prompt is learned using a step-wise distribution alignment strategy. Multiple internal processes may occur between two steps. The style prompt is represented in an image format for better visualization.}
\vspace{-5mm}
\label{overview}
\end{figure*}

In this work, we address the aforementioned challenges by implicitly reducing domain gaps through the optimization of the prediction distribution. Specifically, we propose Step-wise Distribution-aligned Style Prompt Tuning (StepSPT) to tackle SF-CDFSL, as illustrated in Figure~\ref{overview}. In Section~\ref{sec3.1}, we first explain the motivation behind introducing the style prompt and step-wise distribution alignment. We define prompt tuning in Section~\ref{sec3.1.1} and highlight that existing prompts are not suitable for CDFSL. Drawing inspiration from the ability of batch normalization to adjust style and influence data distribution~\cite{yazdanpanah2022visual}, StepSPT incorporates a learnable style prompt that modifies data distribution while constraining the complexity of the hypothesis space. Furthermore, in Section~\ref{sec3.1.2}, we discuss the challenges of optimizing the target prediction distribution through single-step search and provide theoretical analysis explaining how step-wise distribution alignment addresses these challenges. Section~\ref{sec3.2} details the proposed method, including the design of the style prompt in Section~\ref{sec3.2.1}, the dual-phase optimization process in StepSPT in Section~\ref{sec3.2.2}, and a summary of the overall pipeline in Section~\ref{sec3.2.3}. Algorithm~\ref{pipeline} in Section~\ref{sec3.2.3} outlines the pipeline of StepSPT and summarizes its main operational steps. This provides a clear and comprehensive understanding of the framework.

\subsection{Warm-Up: Why Style Prompt and Step-wise Distribution Alignment?}
\label{sec3.1}
\subsubsection{Why Style Prompt?} 
\label{sec3.1.1}
Prompt tuning is typically used when the pretrained model is frozen. Based on the descriptions in \cite{lester2021, liu2023pre}, we define prompt tuning as follows.
\begin{MyDef}
\label{def:pt}
\textbf{Prompt Tuning.} Given a frozen pretrained model $\theta$, a sample space $\mathcal{X}$, and the corresponding label space $\mathcal{Y}$, where each input instance $x \in \mathcal{X}$ and each label $y \in \mathcal{Y}$, the goal of prompt tuning is to use a prompting mechanism $f_{p}(\cdot)$ to modify $x$ into a prompt $x_{p} = f_{p}(x)$, such that the model maximizes the likelihood $P_{\theta}(y|x_{p})$ of the correct label $y$. Here, $f_{p}(\cdot)$ can be designed either manually by humans or automatically through algorithmic methods.
\end{MyDef}
Take the multi-modal classification task using CLIP~\cite{radford2021learning} as an example. For each image $x$, a human-designed prompt~\cite{lester2021,liu2023pre} could be a fixed sentence pattern such as $f_{p}(x)$ = "A photo of a [$class_{x}$]." In contrast, an automatically designed prompt~\cite{jia2022visual,zhou2022learning,zhou2022conditional} is typically learnable parameters like $f_{p}(x)$ = "\textit{[V]$_{1}$ [V]$_{2}$ $\dots$ [V]$_{M}$ [$class_{x}$]}", where all \textit{[V]} are learnable parameters that can be optimized.

\begin{figure}[!t]
  \centering
  \includegraphics[width=0.48\textwidth]{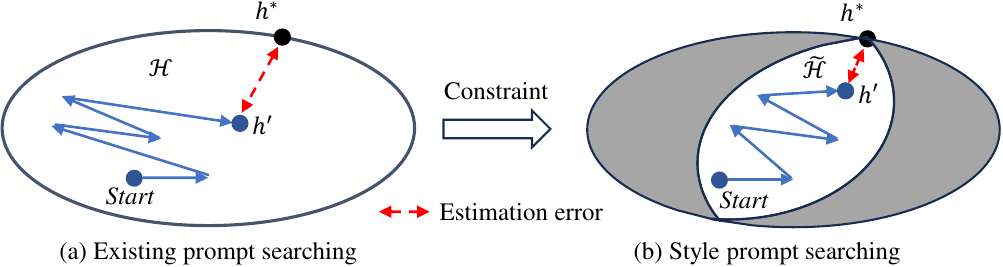}
  \caption{(a) The standard prompt searching. (b) The style prompt search. $h^{*}$ is the expected risk and $h^{'}$ is the empirical risk. $\mathcal{H}$ and $\tilde{\mathcal{H}}$ represent the hypothesis space before and after constraints, respectively. The objective is to minimize the estimation error.}
  \vspace{-5mm}
  \label{sp}
\end{figure}

Existing automatically designed prompts are unsuitable for cross-domain scenarios. Inspired by the influence of style on data distribution~\cite{yazdanpanah2022visual}, this paper proposes a style prompt to adjust the target data distribution and mitigate performance degradation caused by cross-domain issues. Additionally, the style prompt constrains the complexity of the hypothesis space, making it easier to find optimal hypotheses and improving suitability for FSL tasks. As shown in Figure~\ref{sp} (a), existing methods search for optimal prompts in the entire prompt hypothesis space $\mathcal{H}$. Our goal is to minimize the estimation error, which is challenging for the target FSL task. This paper constrains the complexity of $\mathcal{H}$ ($\tilde{\mathcal{H}}$ represents the hypothesis space after constraints) by focusing adjustments solely on the style aspect, as shown in Figure~\ref{sp} (b), which speeds up the searching. Furthermore, inspired by~\cite{wang2020generalizing}, $\tilde{\mathcal{H}}$ is more conducive to FSL tasks.

\subsubsection{Why using Step-wise Distribution Alignment?}
\label{sec3.1.2}
Since the source data is inaccessible in SF-CDFSL, explicitly reducing the domain gap between source and target data becomes infeasible. To address the challenges inherent in SF-CDFSL and optimize the prompts for reducing distribution disparity, this paper proposes an implicit domain alignment strategy. Specifically, let $P_{\theta}$ denote the initial prediction distribution obtained by the pre-trained model $\theta$ on $\mathcal{D}^{t}$. Our aim is to approximate the ideal prediction distribution $P_{t}$, which is unknown and typically markedly distinct from $P_{\theta}$ (\ie exhibiting a significant distribution gap). Therefore, the problem can be framed as a distribution optimization problem to minimize this gap,
\begin{equation}
  arg \min_{\mathcal{\theta}} \mathbf{D} (\hat{P}_{t}, P_{t}),
  \label{eq:all}
\end{equation}
where $\hat{P}_{t}$ represents the predicted distribution by the model, and $\mathbf{D}(\cdot, \cdot)$ denotes the discrepancy between two distributions. However, two key challenges arise in optimizing $\hat{P}_{t}$: (1) the unknown nature of $P_{t}$ complicates optimization efforts, and (2) the substantial domain gap between $P_{\theta}$ and $P_{t}$ makes achieving satisfactory results through a one-step search for $\hat{P}_{t}$ difficult~\cite{kumar2020understanding,tang2023source,abnar2021gradual}.

Inspired by \cite{tang2023source, abnar2021gradual}, this paper transforms the challenging one-step search for $\hat{P}_{t}$ into a multi-step alignment problem by progressively reducing distribution discrepancies. Specifically, we construct a step-wise prediction distribution flow from $P_{\theta}$ to $P_{t}$, expressed as $P_{\theta_{0}} \to P_{\theta_{1}} \to \dots \to P_{\theta_{e}} \to \dots \to P_{\theta_{E}}$, where $P_{\theta_{0}}=P_{\theta}$ and $P_{\theta_{E}}=\hat{P}_{t}$ with $\theta{E}=\theta_{t}$. Here, $\theta_{e}$ represents the $e$-th intermediate model used to estimate $P_{\theta_{e}}$. Our objective is to optimize Eq.~\ref{eq:all}, such that $\hat{P}_{t}=P_{t}$, where $\hat{P}_{t}$ and $P_{t}$ share the same label space $\mathcal{Y}$. Based on Theorem 3.2 in~\cite{kumar2020understanding}:
\begin{MyTheory}
    \textit{Given distribution $P$, $Q$ with distance D($P$,$Q$)=$\tau<\frac{1}{R}$ ($\frac{1}{R}$ stands for the regularization strength of models to be learned) and P($\mathcal{Y}$)=Q($\mathcal{Y}$) (no label shift). Suppose P, Q satisfy the bounded data assumption, and we have initial model $\theta$ with objective loss L($\theta$,P), and $n$ unlabeled samples $S$ from Q, we set $\theta^{'}$=ST($\theta$,S) letting the objective loss L($\theta^{'}$,Q)$<\alpha^{*}$ ($\alpha^{*}$ is a given small loss), then
    \begin{equation}
    \footnotesize
    L(\theta^{'},Q) \le \frac{2}{1-\tau R}L(\theta,P)+\alpha^{*}+O ( \frac{1}{\sqrt{n} }  ).
\end{equation}
}
\end{MyTheory}
It can be observed that $L(\theta^{'},Q)$ decreases as $\tau$ becomes smaller. However, minimizing $\tau$ is challenging when there is a large domain gap between $P$ and $Q$. Inspired by~\cite{tang2023source}, this paper attempts to transform the optimization problem of $\tau$ into multiple smaller problems $\left\{\tau_{i}\right\}^{E}_{i=1}$ ($\tau=\sum^{E}_{i=1}\tau_{i}$). The performance upper bound is presented in the following theorem,
\begin{MyTheory}
    \textit{Given a distribution flow from $P_{\theta_{0}}$ to $P_{\theta_{E}}$, denoted as $P_{\theta_{0}}\to P_{\theta_{1}}\to \dots\to P_{\theta_{E}}$, $P_{\theta_{E}}=\hat{P}_{t}$, and P($\mathcal{Y}$)=Q($\mathcal{Y}$), and the data is bounded. Distribution distances are $\left\{\tau_{i}\right\}^{E}_{i=1}$, and $\tau_{m} = max(\left\{\tau_{i}\right\}^{E}_{i=1})$, then 
\begin{equation}
\footnotesize
\begin{aligned}
    L(\theta_{E},P_{\theta_{E}}) & \le \frac{2}{1-\tau_{E} R}L(\theta_{E-1},P_{\theta_{E-1}})+\alpha^{*}+O ( \frac{1}{\sqrt{n} }  ), \\
    & \le \frac{2}{1-\tau_{m} R}L(\theta_{E-1},P_{\theta_{E-1}})+\alpha^{*}+O ( \frac{1}{\sqrt{n} }   ),
\end{aligned}
\end{equation}
if source model has low loss $\alpha_{0}\ge\alpha^{*}$ on the source domain, according to the formula for the sum of a geometric series,
    \begin{equation}
    \footnotesize
\begin{aligned}
    L(\theta_{E},P_{\theta_{E}}) \le ( \frac{2}{1-\tau_{m} R} )^{E+1}( \alpha_{0}+O( \frac{1}{\sqrt{n} } ) ).
\end{aligned}
\label{pr}
\end{equation}
}
\end{MyTheory}

The tightness of the upper bound in Eq. \ref{pr} depends on both $\frac{2}{1-\tau_{m} R}$ and $E$, with the factor $\frac{2}{1-\tau_{m} R}$ being a critical determinant. To better understand the behavior of growth within the recursive process, we apply a Fourier transform to convert the upper bound of Eq. \ref{pr} into the frequency domain, facilitating a more intuitive analysis of its stability and convergence, as illustrated below:
    \begin{equation}
    \footnotesize
\begin{aligned}
    \hat{L}(\omega) & = \sum_{E=0}^{\infty} ( \frac{2}{1-\tau_{m} R} )^{E+1}( \alpha_{0}+O( \frac{1}{\sqrt{n} } ) ) e^{-j\omega E}  \\
    & = ( \alpha_{0}+O( \frac{1}{\sqrt{n} } ) ) \sum_{E=0}^{\infty} ( \frac{2}{1-\tau_{m} R} )^{E+1} e^{-j\omega E}  \\
    & = ( \alpha_{0}+O( \frac{1}{\sqrt{n} } ) ) ( \frac{2}{1-\tau_{m} R} ) \sum_{E=0}^{\infty} ( \frac{2e^{-jw}}{1-\tau_{m} R} )^{E}
\end{aligned}
\label{ft}
\end{equation}
where $\omega$ denotes the frequency, and $j$ represents the imaginary unit in complex numbers. At this point, we convert the exponential function in Eq. \ref{pr} into a series summation representation in Eq. \ref{ft}. To ensure the upper bound does not diverge, the geometric series in Eq. \ref{ft} must converge, which requires $\left| \frac{2e^{-jw}}{1-\tau_{m} R} \right| < 1$, \ie $\tau_{m} R > 3$. The final frequency domain formula for Eq. \ref{ft} is:
    \begin{equation}
    \footnotesize
\begin{aligned}
    \hat{L}(\omega) & = ( \alpha_{0}+O( \frac{1}{\sqrt{n} } ) ) \frac{2}{(1-\tau_{m} R)-2e^{-j\omega}},  \\
    & \textit{with} \quad \tau_{m} R > 3.
\end{aligned}
\label{ft1}
\end{equation}
Eq.~\ref{ft1} shows that, under weak regularization constraints, the upper bound of $L(\theta_{E},P_{\theta_{E}})$ can be lowered by decreasing $\tau_{m}$ and $\omega$. Here, $\omega$ reflects the oscillation in the recursive growth and depends on $\tau_{m}$, with larger $\tau_{m}$ resulting in greater $\omega$.
By introducing the condition $\tau_{m} R > 3$ into Eq. \ref{pr}, we can also deduce that $\frac{2}{1-\tau_{m}R} < 1$, implying that $( \frac{2}{1-\tau_{m}R} )^{E+1} < 1$. Therefore, $L(\theta_{E},P_{\theta_{E}})$ achieves a tight upper bound. However, $\tau_{m} = \max(\left\{\tau_{i}\right\}^{E}_{i=1})$ can only be obtained at the end of the alignment process. Instead of using $\tau_{m}$, we aim to reduce all $\left\{\tau_{i}\right\}^{E}_{i=1}$ to optimize $L(\theta_{E},P_{\theta_{E}})$. Consequently, a step-wise distribution alignment strategy is proposed to reduce each $\tau_{i}$, converting the original optimization (Eq.~\ref{eq:all}) into the following multiple sub-problems:
\begin{equation}
\begin{aligned}
  arg\min_{\theta}\mathbf{D}\left ( P_{\theta_{e-1}}, P_{\theta_{e}} \right ), \quad e = 1,\dots ,E,
  \label{eq:sub}
\end{aligned}
\end{equation}
the \textit{e}-th sub-problem refers to a single-step search that computes the current prediction distribution $P_{\theta_{e}}$, given the previous prediction distribution $P_{\theta_{e-1}}$. In summary, a step-wise distribution alignment strategy is proposed to transform the challenging one-step prediction distribution optimization problem (Eq. \ref{eq:all}) into a simple multi-step distribution alignment problem (Eq. \ref{eq:sub}). This strategy not only simplifies the optimization process but also enhances the model's adaptability to domain shifts by incrementally reducing domain gaps. As a result, the proposed strategy provides an effective solution for improving cross-domain few-shot learning performance.

\subsection{StepSPT: Step-wise Distribution-aligned Style Prompt Tuning}
\label{sec3.2}
The source model $\theta$ maps image samples to a high-dimensional feature space, denoted as $\theta: \mathcal{X}_{s} \to \mathbb{R}^{dm}$, where \textit{dm} is the feature dimension. Additionally, a classifier $\psi: \mathbb{R}^{dm} \to \mathbb{R}^{N}$ is then applied, forming $\psi \circ \theta$. This section introduces Step-wise Distribution-aligned Style Prompt Tuning (StepSPT), as shown in Figure~\ref{overview}. To minimize learning costs, the parameters of $\theta$ remain fixed. Firstly, StepSPT proposes a style prompt $\omega$ (Section~\ref{sec3.2.1}) to adjust input styles to adapt $\psi \circ \theta$. Additionally, StepSPT explores a dual-phase optimization process (Section~\ref{sec3.2.2}) consisting of external and internal processes. The external process is designed to optimize $\omega$ using a step-wise distribution alignment while keeping $\psi \circ \theta$ frozen. The query set $Q$ is introduced into the external process to bridge domain gaps and to mitigate overfitting. To avoid error accumulation in the step-wise alignment process, this paper introduces a credible group $\mathcal{G}$ to correct alignment errors between steps. Specifically, some samples in $P_{\theta_{e-1}}$ are first selected through entropy sorting and prototype sorting to form $\mathcal{G}$, and then samples in $P_{\theta_{e}}$ are selected by chain-search to pair with samples in $\mathcal{G}$. Finally, by aligning these data pairs, noise and errors can be reduced. The internal process, meanwhile, optimizes $\psi$ using only the labeled support set $S$, while keeping $\theta$ and $\omega$ frozen. There can be one external and multiple internal processes in a step. To illustrate the StepSPT process more clearly, Algorithm~\ref{pipeline} is provided in Section~\ref{sec3.2.3}.

\subsubsection{Style Prompt}
\label{sec3.2.1}
Given that image style significantly influences domain shift~\cite{fu2023styleadv,fu2022wave}, this paper devises a style prompt $\omega$ to interact with input images in $\mathcal{X}$, adjusting their styles to align with the desired distribution and adapt to the source model. Unlike existing approaches that treat prompts and input data as independent entities, this work draws inspiration from batch normalization's ability to modify image styles, designing the learnable $\omega$ to adjust input image styles, as shown in the following formulation:
\begin{equation}
\footnotesize
  x_{p} = f_{p}(x) = \omega_{1} \frac{x-\mu_{\mathcal{X}}}{\sqrt{\sigma^{2}_{\mathcal{X}} +\varepsilon } }+\omega_{2}, \quad\quad\quad  x \in \mathcal{X},
  \label{eq:prompt}
\end{equation}
where $\omega_{1}$ and $\omega_{2}$ (collectively referred to as $\omega$) are the learnable parameters of style prompt, $\mathcal{X}=\left \{\mathcal{X}_{s}, \mathcal{X}_{q}\right \}$, $\mu_{\mathcal{X}}$ and $\sigma^{2}_{\mathcal{X}}$ are the mean and variance of $\mathcal{X}$,
\begin{equation}
\footnotesize
  \mu_{\mathcal{X}} = \frac{1}{m}  {\textstyle \sum_{i=1}^{m}} x_{i},  \quad\quad\quad
  \sigma^{2}_{\mathcal{X}} = \frac{1}{m}  {\textstyle \sum_{i=1}^{m}} (x_{i}-\mu_{\mathcal{X}})^{2},
  \label{eq:mean}
\end{equation}
where $m$ is the sample number of $\mathcal{X}$, the obtained $x_{p}$ is the adjusted $x$ through $\omega$, where $\omega_{1}$ and $\omega_{2}$ are updated by a step-wise distribution alignment strategy proposed in the external process, which will be introduced in Section~\ref{sec3.2.2}.

\subsubsection{Dual-phase Optimization Process}
\label{sec3.2.2}
The dual-phase optimization process consists of external and internal processes. The external process focuses on optimizing $\omega$ through the proposed step-wise distribution alignment strategy, aiming to implicitly reduce the distance between the source and target domains by aligning their distributions. Meanwhile, the internal process updates $\psi$ to adapt to the target domain and improve classification accuracy.

\textbf{External Process.}
In the external process, $\omega$ is updated while keeping $\theta$ and $\psi$ frozen. Since the Wasserstein distance is beneficial for solving cross-domain problems~\cite{shen2018wasserstein}, we employ it to quantify distribution shift~\cite{kumar2020understanding}, denoted as $\mathbf{D}_{wass}(\cdot, \cdot)$. For any adjacent distributions, the shift measure for $N$-way classification is expressed as:
\begin{equation}
\footnotesize
\begin{aligned}
  & \mathbf{D} _{wass} \left ( P_{\omega_{e-1}},P_{\omega_{e}} \right ) = \max\left \{ d_{0},\dots , d_{n},\dots , d_{N-1} \right \}, \\
  &  d_{n} = \textbf{W}_{\infty}\left ( P_{\omega_{e-1}} ( x^{e-1}|y^{e-1}=n ), P_{\omega_{e}} ( x^{e}|y^{e}=n )  \right ),
  \label{eq:wass}
\end{aligned}
\end{equation}
where $\textbf{W}_{\infty}(\cdot,\cdot)$ denotes the Wasserstein-infinity distance, $x^{e-1}$ and $x^{e}$ represent samples satisfying $P{\omega_{e-1}}$ and $P_{\omega_{e}}$, respectively, and $y^{e-1}$ and $y^{e}$ denote their corresponding labels. The conditional distributions $P_{\omega_{e-1}} ( x^{e-1}|y^{e-1}=n )$ and $P_{\omega_{e}} ( x^{e}|y^{e}=n )$ represent the probability distributions for the $n$-th category according to $P_{\omega_{e-1}}$ and $P_{\omega_{e}}$, respectively. For each of the $N$ categories, we calculate the Wasserstein-infinity distances and use the largest distance to quantify the distribution shift. However, in the query set $Q$, exact category labels are unavailable. To overcome this challenge, we aim to minimize the distances $\left\{ d_{0},\dots , d_{n},\dots , d_{N-1} \right\}$ without depending on category labels, as described by the following equation:
\begin{equation}
\footnotesize
\begin{aligned}
  &  d_{e} = \textbf{W}_{\infty}\left ( P_{\omega_{e-1}} ( x^{e-1}|y^{e-1} ), P_{\omega_{e}} ( x^{e}|y^{e} )  \right ), \\
  &  with  \quad\quad\quad\quad y^{e-1}=y^{e},
  \label{eq:de}
\end{aligned}
\end{equation}
Since $P_{\omega_{e-1}}$ is not aligned with $P_{t}$, which means not all samples in $P_{\omega_{e-1}}$ are credible. Therefore, addressing the minimization of $d_{e}$ involves two key challenges: (1) determining the credible data pairs supporting $d_{e}$ estimation, and (2) reducing $d_{e}$ using these data pairs. 

The proposed step-wise distribution alignment strategy addresses these challenges through two components: (1) the distribution gap estimation and (2) the distribution alignment. 

\begin{figure}[t]
  \centering
  \includegraphics[height=1.2\linewidth]{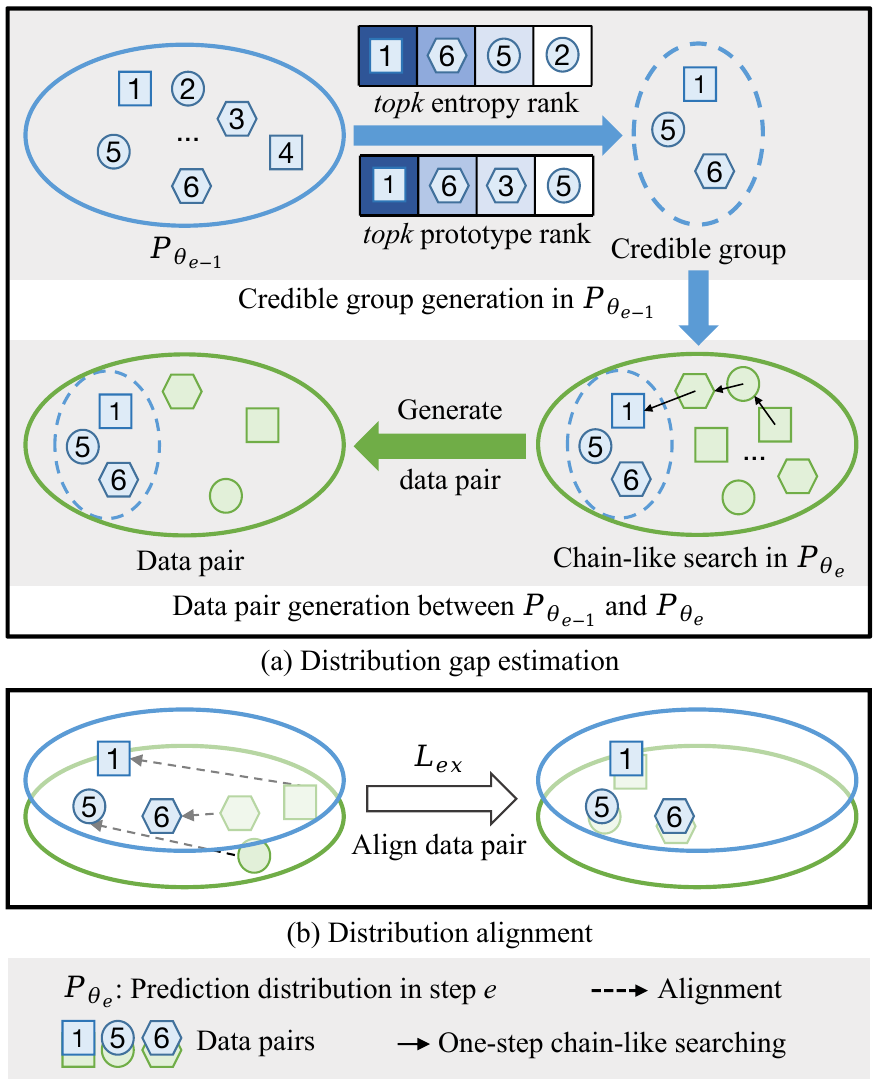}
  \vspace{-2mm}
  \caption{The overview of the Step-wise distribution alignment strategy. The goal is to align the examples from $P_{\theta_{e}}$ to those from $P_{\theta_{e-1}}$. This is achieved through two stages: (a) Distribution gap estimation and (b) Distribution alignment. The color gradient from dark to light represents the ranking of choices from best to worst.
  }
  \vspace{-4mm}
  \label{stepp}
\end{figure}

(1) \textit{Distribution Gap Estimation.}
In the distribution gap estimation phase, as the distribution in $P_{\theta_{e-1}}$ may not align with the optimal target distribution $P_{t}$, the initial task in distribution gap estimation is to establish a credible group $\mathcal{G}$ within $P_{\theta_{e-1}}$ as a trusted distribution label, as shown in the ``Credible group generation in $P_{\theta_{e-1}}$'' section of Figure~\ref{stepp} (a). This entails ensuring that samples exhibit high classification confidence in specific categories through self-supervised or unsupervised methods. Data from $P_{\theta_{e}}$ is then matched with $\mathcal{G}$ to form credible data pairs. Two methods are employed for generating $\mathcal{G}$: entropy-based ranking~\cite{liu2021source,yang2020unsupervised} and prototype-based ranking~\cite{snell2017prototypical}.

Initially, we generate the entropy-based ranking credible group $\mathcal{G}_{en}$ as follows:
\begin{equation}
\footnotesize
\begin{aligned}
  \mathcal{G}_{en}=\left\{ \textbf{p}_{i} | \textbf{p}_{i} \in P_{\theta_{e-1}}, i \in \textit{topk}(H,\alpha \cdot m) \right\},
  \label{eq:enn}
\end{aligned}
\end{equation}
where $H=\left\{h_{i}\right\}^{m}_{i=1}$ is an entropy set, with $h_{i}=-\sum \textbf{p}_{i} \text{log} \textbf{p}_{i}$, $\textbf{p}_{i}=\psi\circ \theta \circ \omega(x_{i})$ and $x_{i} \in \mathcal{X}$, while $m$ denotes the number of samples, and $\textit{topk}(H,\alpha \cdot m)$ selects the top $\alpha \cdot m$ lowest elements from $H$. 

\begin{figure}[t]
  \centering
  \includegraphics[height=4.4cm]{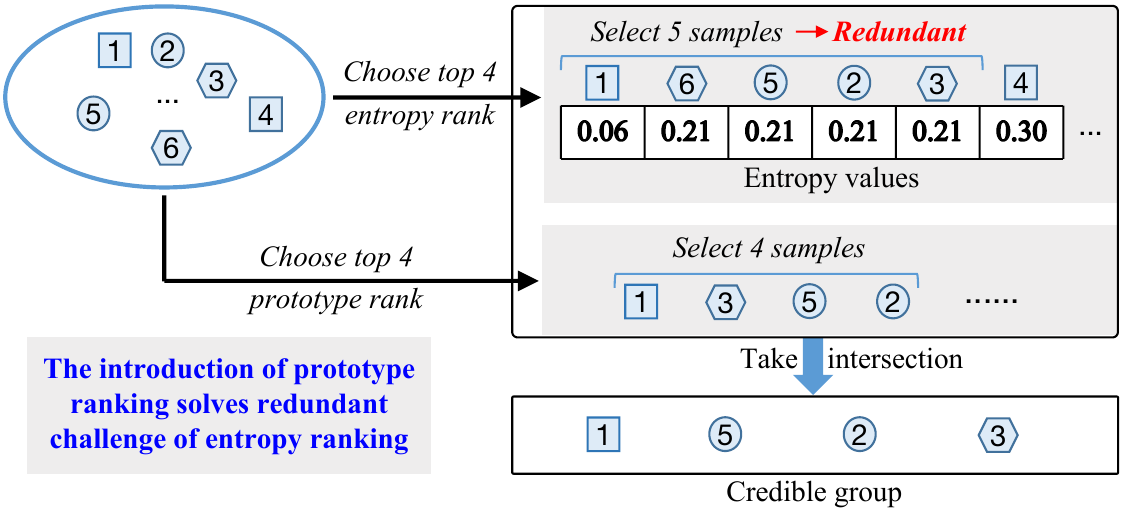}
  \vspace{-0.3cm}
  \caption{Problem of only using $\mathcal{G}_{en}$. Samples 6, 5, 2, 3 have the same entropy values, the top 4 selection of $\mathcal{G}_{en}$ become redundant and ambiguous. The introduction of prototype ranking $\mathcal{G}_{pro}$ can address this challenge.}
  \vspace{-5mm}
  \label{why}
\end{figure}

However, as shown in Figure~\ref{why}, the entropy-based group $\mathcal{G}_{en}$ has a many-to-one problem of prediction distributions and entropy values, making $\mathcal{G}_{en}$ redundant and ambiguous. For example, in Figure~\ref{why}, we should choose the 4 samples with the highest entropy values. However, the samples 6, 5, 2, and 3 have the same entropy value, which ties them for second place (\ie the same entropy value, \eg 0.21, corresponds to multiple samples, \eg 6, 5, 2, 3). One of them needs to be excluded, but no indicator determines which sample to exclude.
We alleviate this problem from the class-aware dimension. Specifically, a prototype-based ranking credible group $\mathcal{G}_{pro}$ is introduced as follows,
\begin{equation}
\footnotesize
\begin{aligned}
  \mathcal{G}_{pro}=\left\{ \textbf{p}_{i} | \textbf{p}_{i} \in P_{\theta_{e-1}}, i \in \textit{topk}(A,\gamma \cdot m) \right\},
  \label{eq:go}
\end{aligned}
\end{equation}
where $\textit{topk} (A,\gamma \cdot m)$ means select the top $\gamma\cdot m$ samples from $A=\left\{ a_{i} \right\}^{m}_{i=1}$ over the target data, $a_{i}=min(\textbf{v}_{i})$, where $\textbf{v}_{i}=\left\{ v_{i,1}, v_{i,2}, ..., v_{i,N} \right\}$, with $v_{i,n}$ is the cosine distance between the feature vector $\textbf{p}_{i}$ and the prototype $\textbf{o}_{n}$, calculated as follows:
\begin{equation}
\footnotesize
\begin{aligned}
  v_{i,n} = cos(\textbf{p}_{i}, \textbf{o}_{n})=\frac{\textbf{p}_{i} \cdot \textbf{o}_{n}}{\| \textbf{p}_{i} \| \| \textbf{o}_{n} \|},
  \label{eq:vin}
\end{aligned}
\end{equation}
and $\textbf{o}_{n}$ is obtained by,
\begin{equation}
\footnotesize
\begin{aligned}
  \textbf{o}_{n} = \frac{1}{K} \sum_{k=1}^{K}\psi \circ \theta \circ \omega(S),
  \label{eq:on}
\end{aligned}
\end{equation}

Finally, the credible group $\mathcal{G}$ is obtained by an intersection operation as $\mathcal{G}=\mathcal{G}_{pro}  \cap \mathcal{G}_{en}$.

We then match the examples in $P_{\theta_{e}}$ with $\mathcal{G}$. To construct the data pairs between $P_{\theta}^{e}$ and $P_{\theta}^{e-1}$, the credible neighbor $\textbf{p}'_{i}$ from $\mathcal{G}$ is identified through an iterative chain-based search for $\textbf{p}_{i}$ (from $P_{\theta_{e}}$). Specifically, beginning with $\textbf{p}_{i}$, we perform a one-step search, based on cosine distance, to find its nearest neighbor $\hat{\textbf{p}}_{i}^{1}$. If $\hat{\textbf{p}}_{i}^{1}$ does not belong to $\mathcal{G}$, we perform another search. This time, the search begins with $\hat{\textbf{p}}_{i}^{1}$ to find a new nearest neighbor $\hat{\textbf{p}}_{i}^{2}$. This process continues iteratively until we reach $\mathcal{G}$, resulting in a search flow ${ \textbf{p}_{i} \to \hat{\textbf{p}}_{i}^{1} \to \hat{\textbf{p}}_{i}^{2} \to ... \to \textbf{p}'_{i} }$. Here, $\textbf{p}'_{i} \in \mathcal{G}$ serves as the data pair of $\textbf{p}_{i}$, as illustrated in the "Data pair generation between $P_{\theta_{e-1}}$ and $P_{\theta_{e}}$" section of Figure~\ref{stepp} (a).

(2) \textit{Distribution Alignment.}
During distribution alignment, we mitigate distribution shift by aligning these data pairs while maintaining category consistency. Building on the theoretical link between mutual information and pairwise losses \cite{boudiaf2020unifying}, we use the following objective to minimize distribution shift,
\begin{equation}
\footnotesize
\begin{aligned}
  L_{MI}(\omega_{e})=\min_{\omega_{e}}\left [ -I\left ( \textbf{p}_{i}, \textbf{p}'_{i} \right ) \right ],
  \label{eq:im}
\end{aligned}
\end{equation}
Here, $I(\cdot,\cdot)$ denotes the mutual information function \cite{paninski2003estimation}, computed in a manner similar to \cite{ji2019invariant}. Given that limited data is insufficient to adequately represent the probability distribution, optimizing solely based on $L_{MI}$ is limited in effectiveness. To address the issue of limited data representation, we introduce a diversity loss to promote balanced category predictions \cite{jabi2019deep}.
\begin{equation}
\footnotesize
\begin{aligned}
  L_{KL}(\omega_{e})=\min_{\omega_{e}}\sum_{n=1}^{N}KL( \bar{\textbf{p}}_{n}, \frac{1}{N} ).
  \label{eq:kl}
\end{aligned}
\end{equation}
Here, $KL(\cdot,\cdot)$ denotes the KL-divergence loss function, and $\bar{\textbf{p}}_{n}=\frac{1}{m}\sum_{i=1}^{m}\textbf{p}_{i,n}$ represents the empirical label distribution, where $\textbf{p}_{i,n}$ is the probability of $x_{i}$ in the $n$-th category. The final objective in the external process is:
\begin{equation}
\footnotesize
\begin{aligned}
  L_{ex}(\omega_{e})=L_{MI}(\omega_{e}) + \sigma L_{KL}(\omega_{e}),
  \label{eq:external}
\end{aligned}
\end{equation}
where $\sigma$ is the weighting factor for $L_{KL}$. The distribution alignment is illustrated in Figure~\ref{stepp} (b). Generally, the external process adjusts the style prompt $\omega$ through step-wise distribution alignment, allowing the model to progressively adapt to domain shifts and maintain robustness across data distributions.

\textbf{Internal Process.}
Following the style prompt updates, the classifier $\psi$ is optimized using cross-entropy loss solely with $S$ in the internal process, keeping $\theta \circ \omega$ frozen. The cross-entropy loss is as follows:
\begin{equation}
\footnotesize
\begin{aligned}
  L_{in}(\psi_{e})=CE(\psi_{e}\circ \theta|\omega_{e-1}(x_{i}))=-\sum_{i=1}^{N \times K}y_{i} \text{log} \textbf{p}_{i}.
  \label{eq:ce}
\end{aligned}
\end{equation}
where $\textbf{p}_{i}=\theta \circ \omega_{e-1}(x_{i})$, $\left\{x_{i}, y_{i}\right\} \subset S$, and $N \times K$ means the number of support set $S$. There can be multiple internal processes in a step, which shows the iterative nature of the optimization and emphasizes the importance of refining the classifier $\psi$ to achieve stable and accurate predictions.

\subsubsection{Overall Pipeline of StepSPT}
\label{sec3.2.3}
\begin{algorithm}[!t]
\caption{StepSPT process.}
\begin{algorithmic}
\STATE
\STATE {\textbf{Input:}} Support set $(\mathcal{X}_s,\mathcal{Y}_s)$, query set $\mathcal{X}_q$, initialized style prompt $\omega_{0}$, frozen encoder $\theta$, and initialized classifier $\psi_{0}$.

\STATE {\textbf{Output:}} Optimized style prompt $\omega$ and classifier $\psi$.


\STATE $\textbf{p}_{i}=\psi_{0} \circ \theta \circ \omega_{0}(x_{i})$, \quad\quad  $x_{i} \in \left\{\mathcal{X}_s, \mathcal{X}_q\right\}$, $\textbf{p}_{i} \in P_{\omega_{0}}$

\STATE {\textbf{For} \textit{step=1,\dots,e,\dots ,E} \textbf{do}}

\STATE \hspace{0.5cm} Calculate the entropy $h_{i}$ and prototype centers $\textbf{o}_{n}$ in $P_{\omega_{e-1}}$

\STATE \hspace{0.5cm} Get $a_{i}$ and $\textbf{o}_{n}$ in $P_{\omega_{e-1}}$ 

\STATE \hspace{0.5cm}  Get $H=\left\{h_{i}\right\}^{m}_{i=1}$ and $A=\left\{ a_{i} \right\}^{m}_{i=1}$

\STATE \hspace{0.5cm} Get $\mathcal{G}_{en}=\left\{ \textbf{p}_{i} | \textbf{p}_{i} \in P_{\omega_{e-1}}, i \in \textit{topk}(H,\alpha \cdot m) \right\}$

\STATE \hspace{0.5cm} Get $\mathcal{G}_{pro}=\left\{ \textbf{p}_{i} | \textbf{p}_{i} \in P_{\omega_{e-1}}, i \in \textit{topk}(A,\gamma \cdot m) \right\}$

\STATE \hspace{0.5cm} Get the final credible group $\mathcal{G}=\mathcal{G}_{pro}  \cap \mathcal{G}_{en}$



\STATE \textcolor{blue}{\hspace{0.5cm} \textbf{for} \textit{epoch=1,\dots, max epochs/E} \textbf{do}} 

\STATE \textcolor{blue}{\hspace{1.0cm} \textbf{for} each mini-batch \{$x_i$,$y_i$\} in $(\mathcal{X}_s,\mathcal{Y}_s)$ \textbf{do}}

\STATE \textcolor{blue}{\hspace{1.5cm} $\textbf{p}_{i}=\psi_{e-1} \circ \theta \circ \omega_{e-1}(x_{i})$,}

\STATE \textcolor{blue}{\hspace{1.5cm} \parbox[t]{\dimexpr0.9\linewidth-\algorithmicindent}{Compute Eq.~\ref{eq:ce} and update $\psi_{e-1}$ as $\psi_{e}$;}}

\STATE \textcolor{blue}{\hspace{1.0cm} $\textbf{end for}$}

\STATE \textcolor{blue}{\hspace{0.5cm} $\textbf{end for}$}

\STATE \hspace{0.5cm} $\textbf{p}_{i}=\psi_{e} \circ \theta \circ \omega_{e-1}(x_{i})$, \quad \quad   $x_{i} \in \left\{\mathcal{X}_s, \mathcal{X}_q\right\}$

\STATE \hspace{0.5cm} Search data pairs $\textbf{p}'_{i}$ in $\mathcal{G}$ for $\textbf{p}_{i}$ through a chain-like search

\STATE \hspace{0.5cm} Compute Eq.~\ref{eq:external} with ($\textbf{p}_{i}$,$\textbf{p}'_{i}$) and update $\omega_{e-1}$ as $\omega_{e}$

\STATE \hspace{0.5cm} $\omega_{e-1}=\omega_{e}$, $\psi_{e-1}=\psi_{e}$, $P_{\omega_{e-1}}=P_{\omega_{e}}$

\STATE {\textbf{end for}}

\end{algorithmic}
\label{pipeline}
\end{algorithm}
Algorithm~\ref{pipeline} presents the overall pipeline of StepSPT, highlighting both the internal and external processes. Initially, the support set ($\mathcal{X}_{s}$, $\mathcal{Y}_{s}$) and query set $\mathcal{X}_{q}$ are provided as input, along with the style prompt $\omega_{0}$ and classifier $\psi_{0}$, which are randomly initialized. The encoder $\omega$ remains frozen, having been pre-trained on source data to ensure stable feature extraction. The initial prediction $\textbf{p}_{i}=\psi_{0} \circ \theta \circ \omega_{0}(x_{i})$ is calculated for each $x_{i} \in \left\{\mathcal{X}_s, \mathcal{X}_q\right\}$. This step helps identify credible samples through entropy $h{i}$ and prototype centers $\textbf{o}_{n}$, which are then used to form the credible group $\mathcal{G}$. The internal process (highlighted in \textcolor{blue}{blue} in Algorithm~\ref{pipeline}) iteratively optimizes the classifier $\psi$ using cross-entropy loss for a specified number of epochs, \textit{max epochs} divided by $E$. This step is essential for refining the model's performance on the support set. After the internal process, the external process utilizes the updated classifier $\psi$ and credible group $\mathcal{G}$ to optimize the style prompt $\omega$ through the step-wise distribution alignment strategy. This strategy helps the model better align with the target domain, ensuring that the learned representations remain robust across data distributions. By alternating between the internal and external processes, StepSPT progressively improves the style prompt and classifier, ultimately achieving improved CDFSL performance.

\section{Experiments}
In this section, we introduce the datasets and implementation details. We then present the results of StepSPT, including comparisons with prompt tuning-based methods and state-of-the-art methods (SOTAs), the ablation study, and the extra performance analysis. 

\subsection{Implementation}

\subsubsection{Datasets}
StepSPT is evaluated across four datasets from the BSCD-FSL benchmark~\cite{guo2020broader} and PatternNet~\cite{pattern}. Within the BSCD-FSL scope, domain similarity to scene imagery progressively diminishes from CropDisease to ChestX, ranging from natural environments to medical images. Furthermore, the lower resolution of EuroSAT samples highlights a potential concern: the performance of EuroSAT could be enhanced by increasing image resolution, instead of effectively addressing the cross-domain challenge. To demonstrate that the proposed method effectively addresses the cross-domain problem, this paper includes the high-resolution PatternNet dataset in the evaluation, showcasing StepSPT's strong performance on high-resolution remote sensing data. PatternNet is a large-scale high-resolution remote sensing dataset collected for remote sensing image retrieval. It consists of 38 classes, each containing 800 images with a resolution of $256\times 256$ pixels.

\begin{table*}[!t]
  \caption{Comparison between prompt-tuning based methods and StepSPT in 5-way 1-shot and 5-way 5-shot tasks. \textbf{Backbone} indicates the utilized backbone in the method. Cells highlighted in blue indicate the best performance achieved with each respective backbone.
  }
  \vspace{-2mm}
  \label{compare5W1S}
  \scriptsize 
  \centering
  \setlength{\tabcolsep}{4.5mm}{
  \begin{tabular}{c|lc|cccccc}
    \hline
    Task & \textbf{Methods} & \textbf{Backbone} & \textbf{CropDisease} & \textbf{EuroSAT} & \textbf{ISIC} & \textbf{ChestX} & \textbf{PatternNet} & \textbf{Avg}   \\
    \hline
     \multirow{14}*{\textbf{\rotatebox{270}{5-way 1-shot}}} & Finetuning~\cite{guo2020broader} & ResNet10 & 65.70$\pm$0.85 & 63.44$\pm$0.83 & 33.67$\pm$0.60 &  22.70$\pm$0.41 &  76.06$\pm$0.89 & 52.38    \\ 
     & IM-DCL~\cite{xu2024enhancing} & ResNet10 & 74.23$\pm$0.91 & \cellcolor{blue!25}\textbf{69.91$\pm$0.92} & 34.23$\pm$0.65 &  \cellcolor{blue!25}\textbf{22.89$\pm$0.43} &  76.68$\pm$0.87 & 55.58    \\ 
     & StepSPT (Ours) & ResNet10 & \cellcolor{blue!25} \textbf{78.35$\pm$0.53} & 68.21$\pm$0.42 & \cellcolor{blue!25}\textbf{35.11$\pm$0.63} & 22.83$\pm$1.03 & \cellcolor{blue!25}\textbf{77.83$\pm$0.95} & \cellcolor{blue!25}\textbf{56.47}   \\  
     \cline{2-9}
     & ViT~\cite{dosovitskiy2020image} & ViT & 78.39$\pm$0.88 & 58.60$\pm$0.78 & 30.74$\pm$0.54 &  21.90$\pm$0.91 & 68.13$\pm$0.91 &  51.55   \\
     & VPT~\cite{jia2022visual} & ViT & 75.00$\pm$0.24 & 67.53$\pm$0.59 & 30.80$\pm$0.96 & 20.56$\pm$0.64 & 79.87$\pm$0.69 & 54.75  \\   
     & LoRA~\cite{hu2022lora} & ViT & 81.95$\pm$0.88 & 52.99$\pm$0.80 & \cellcolor{blue!25}\textbf{32.59$\pm$0.32} & 21.59$\pm$0.94 & 74.85$\pm$0.32 & 52.79  \\
     & StepSPT (Ours) & ViT & \cellcolor{blue!25}\textbf{82.56$\pm$0.27} & \cellcolor{blue!25}\textbf{68.47$\pm$0.22} & 32.03$\pm$0.42 &  \cellcolor{blue!25}\textbf{23.55$\pm$1.06} & \cellcolor{blue!25}\textbf{80.84$\pm$0.31} &  \cellcolor{blue!25}\textbf{57.49}  \\
     \cline{2-9}
     & CLIP~\cite{radford2021learning} & CLIP &  78.93$\pm$0.83 & 66.88$\pm$0.80 & 30.45$\pm$0.55 & 21.63$\pm$0.39 & 91.71$\pm$0.55 & 57.92  \\
     & CoOp~\cite{zhou2022learning} & CLIP & 75.25$\pm$1.60 & 70.00$\pm$1.15 & 30.45$\pm$0.59 & 20.91$\pm$0.31 & 90.24$\pm$0.57 & 57.37   \\
     & CoCoOp~\cite{zhou2022conditional} & CLIP & 75.37$\pm$0.78 & \cellcolor{blue!25}\textbf{71.56$\pm$0.85} & 30.07$\pm$0.36 & 20.23$\pm$0.84 & 90.55$\pm$0.31 & 57.56   \\
     & LoRA~\cite{hu2022lora} & CLIP & \cellcolor{blue!25}\textbf{85.75$\pm$0.97} & 71.43$\pm$0.88 & \cellcolor{blue!25}\textbf{33.29$\pm$0.55} & 21.92$\pm$0.91 & \cellcolor{blue!25}\textbf{96.61$\pm$0.90} & \cellcolor{blue!25}\textbf{61.80}  \\
     & StepSPT (Ours) & CLIP & 84.84$\pm$0.72 & 70.01$\pm$0.21 & 32.97$\pm$0.27 & \cellcolor{blue!25}\textbf{22.84$\pm$0.95} & 95.16$\pm$0.51 & 61.16    \\
     \cline{2-9}
     & ConvNeXt~\cite{liu2022convnet} & ConvNeXt & 91.35$\pm$0.64 & 66.25$\pm$0.77 & 33.13$\pm$0.53 & 22.11$\pm$0.55 & 91.93$\pm$0.25 & 60.95  \\
     & LoRA~\cite{hu2022lora} & ConvNeXt & 94.59$\pm$0.38 & 72.53$\pm$0.25 & 36.60$\pm$0.67 & 22.45$\pm$1.03 & 95.40$\pm$0.21 & 64.31  \\
     & StepSPT (Ours) & ConvNeXt & \cellcolor{blue!25}\textbf{95.39$\pm$0.22} & \cellcolor{blue!25}\textbf{73.83$\pm$0.80} & \cellcolor{blue!25}\textbf{37.16$\pm$0.68} & \cellcolor{blue!25}\textbf{23.73$\pm$0.98} & \cellcolor{blue!25}\textbf{96.44$\pm$0.42} & \cellcolor{blue!25}\textbf{65.31}     \\
    \hline
    \hline
    \multirow{14}*{\textbf{\rotatebox{270}{5-way 5-shot}}} & Finetuning~\cite{guo2020broader} & ResNet10 & 88.16$\pm$0.55 & 81.65$\pm$0.63 & 45.87$\pm$0.64 &  25.77$\pm$0.43 &  92.86$\pm$0.39 & 66.86    \\ 
     & IM-DCL~\cite{xu2024enhancing} & ResNet10 & 88.68$\pm$0.58 & 82.90$\pm$0.85 & 45.58$\pm$0.83 &  25.98$\pm$0.47 & 93.03$\pm$0.58 & 67.23    \\ 
    & StepSPT (Ours) & ResNet10 & \cellcolor{blue!25}\textbf{88.84$\pm$0.41} & \cellcolor{blue!25}\textbf{82.91$\pm$0.32} & \cellcolor{blue!25}\textbf{46.41$\pm$0.75} &  \cellcolor{blue!25}\textbf{27.56$\pm$0.90} &  \cellcolor{blue!25}\textbf{93.35$\pm$1.03} & \cellcolor{blue!25}\textbf{67.81}    \\
    \cline{2-9}
    & ViT~\cite{dosovitskiy2020image} & ViT & 95.27$\pm$0.37 & 77.99$\pm$0.57 & 47.66$\pm$0.59 & 25.74$\pm$0.41 & 92.77$\pm$0.36 & 67.89   \\
    & VPT~\cite{jia2022visual} & ViT & 95.20$\pm$0.99 & 78.05$\pm$0.39 & 50.00$\pm$0.44 & 26.63$\pm$1.03 & 93.24$\pm$0.74 & 68.62  \\   
    & LoRA~\cite{hu2022lora} & ViT & 96.65$\pm$0.70 & 78.87$\pm$0.34 & 47.71$\pm$0.54 & 26.05$\pm$0.20 & 93.63$\pm$0.83 & 68.58  \\
    & StepSPT (Ours) & ViT & \cellcolor{blue!25}\textbf{97.07$\pm$0.75} & \cellcolor{blue!25}\textbf{84.88$\pm$0.28} & \cellcolor{blue!25}\textbf{50.39$\pm$0.34} &  \cellcolor{blue!25}\textbf{26.65$\pm$0.11} & \cellcolor{blue!25}\textbf{95.96$\pm$0.65} & \cellcolor{blue!25}\textbf{70.99}   \\
    \cline{2-9}
    & CLIP~\cite{radford2021learning} & CLIP & 92.53$\pm$0.35 & 86.49$\pm$0.82 & 48.67$\pm$0.44 & 25.16$\pm$0.35 & 98.58$\pm$0.59 & 70.29  \\
    & CoOp~\cite{zhou2022learning} & CLIP & 89.26$\pm$1.60 & 85.97$\pm$0.95 & 46.40$\pm$0.85 & 25.37$\pm$0.94 & 98.08$\pm$0.58 & 69.02   \\
    & CoCoOp~\cite{zhou2022conditional} & CLIP & 90.20$\pm$0.34 & 86.07$\pm$0.62 & 46.99$\pm$0.71 & 26.15$\pm$0.83 & 98.71$\pm$0.40 & 69.62   \\
    & LoRA~\cite{hu2022lora} & CLIP & \cellcolor{blue!25}\textbf{96.17$\pm$0.94} & 87.57$\pm$1.04 & 47.91$\pm$0.49 & 23.35$\pm$0.95 & \cellcolor{blue!25}\textbf{99.23$\pm$0.27} & 70.85  \\
    & StepSPT (Ours) & CLIP & 96.01$\pm$0.88 & \cellcolor{blue!25}\textbf{89.40$\pm$1.05} & \cellcolor{blue!25}\textbf{52.12$\pm$0.36} & \cellcolor{blue!25}\textbf{26.36$\pm$0.97} & 99.04$\pm$0.31 & \cellcolor{blue!25}\textbf{72.58}   \\
    \cline{2-9}
    & ConvNeXt~\cite{liu2022convnet} & ConvNeXt & 95.15$\pm$0.29 & 88.11$\pm$0.90 & 52.51$\pm$0.48 & 26.57$\pm$0.21 & 98.52$\pm$0.36 & 72.17  \\
    & LoRA~\cite{hu2022lora} & ConvNeXt & \cellcolor{blue!25}\textbf{97.96$\pm$0.46} & 90.63$\pm$0.84 & 53.21$\pm$0.58 & 26.35$\pm$0.19 & \cellcolor{blue!25}\textbf{99.69$\pm$0.14} & 73.57  \\
    & StepSPT (Ours) & ConvNeXt & 97.11$\pm$0.60 & \cellcolor{blue!25}\textbf{91.07$\pm$0.73} & \cellcolor{blue!25}\textbf{53.99$\pm$0.44} & \cellcolor{blue!25}\textbf{27.11$\pm$0.75} & 99.24$\pm$0.32 & \cellcolor{blue!25}\textbf{73.79}     \\
    \hline
  \end{tabular}
  \vspace{-5mm}
  }
\end{table*}

\subsubsection{Implementation Details}
We average the StepSPT evaluation results for 600 episodes, where each episode contains 100 epochs. We show the performance with average accuracy and 95\% confidence interval. Besides, We used NVIDIA A100 GPUs, each with 40GB of memory. Each experiment was run on a single GPU. On average, each training run of 600 episodes (with ConvNeXt backbone) took approximately 6 hours for 1-shot tasks and 16 hours for 5-shot tasks, respectively. 
A linear classifier is utilized for classification. We utilize Stochastic Gradient Descent (SGD) with a learning rate of 0.01, momentum of 0.9, and weight decay of $10^{-3}$ for updating the prompt and classifier. $\alpha$ and $\gamma$ are set to 0.7 and 0.4, respectively, with $\sigma$ set to 2. To enhance outcomes further, we incorporate label propagation (LP) into our method. Furthermore, StepSPT is evaluated on both 5-way 1-shot and 5-way 5-shot tasks.

\subsection{Compare with prompt tuning-based methods}
\label{prompt_methods}
Our evaluation of StepSPT begins by comparing StepSPT to the baseline and existing prompt-based methods. To ensure objective assessments, we first present the baselines on the backbones of ResNet10, ViT~\cite{dosovitskiy2020image}, CLIP~\cite{radford2021learning}, and ConvNeXt~\cite{liu2022convnet}. All baselines keep the backbone parameters fixed. We use the image encoder in CLIP exclusively, followed by a linear classifier, to evaluate its CDFSL performance. Then we compare StepSPT with prompt-based methods such as VPT~\cite{jia2022visual}, CoOp~\cite{zhou2022learning}, CoCoOp~\cite{zhou2022conditional}, and LoRA~\cite{hu2022lora}. Finally, we present the performance of StepSPT on different pretrained backbones, including ViT-B/16~\footnote{\url{https://github.com/google-research/vision\_transformer}}, CLIP with ViT-B/32~\footnote{\url{https://github.com/openai/CLIP}}, and ConvNeXt-XL~\footnote{\url{https://github.com/huggingface/pytorch-image-models/tree/main/timm}}. The outcomes for both 5-way 1-shot and 5-way 5-shot tasks are detailed in Table~\ref{compare5W1S}.

As shown in Table~\ref{compare5W1S}, in the 5-way 1-shot task, the average performance across the four baselines, ranked in descending order, is as follows: ConvNeXt $\ge$ CLIP $\ge$ ResNet10 $\ge$ ViT. In comparison to near domain tasks like CropDisease, EuroSAT, and PatternNet, both ResNet10 and ConvNeXt perform better than ViT and CLIP in distant domains like ISIC and ChestX. For example, the ResNet10 and ConvNeXt baselines achieve 33.67\% and 33.13\% on the ISIC dataset, outperforming ViT and CLIP, which achieve 30.74\% and 30.45\%, respectively. Furthermore, the performance of ResNet10 and ConvNeXt baselines on ChestX (22.70\% and 22.11\%) is higher than that of ViT and CLIP baselines on ChestX (21.90\% and 21.63\%). This shows that compared to Transformer-based backbones (ViT and CLIP), CNN-based backbones (ResNet10 and ConvNeXt) demonstrate greater advantages in handling cross-domain tasks. This trend is also observable in the current prompt tuning-based methods, including StepSPT. The strong transferability of local features makes them particularly critical for cross-domain tasks, contributing to the superior performance of CNN-based backbones in these scenarios. Additionally, we observe that for near domain tasks such as CropDisease, EuroSAT, and PatternNet, while methods based on ViT (78.39\% for CropDisease) and CLIP (78.93\% for CropDisease) outperform those using ResNet10 (75.70\% for CropDisease), they still fall short compared to approaches leveraging ConvNeXt (91.35\% for CropDisease). This can be attributed to the ability of ConvNeXt to combine the strengths of CNNs~\cite{liu2022convnet,su2024boosting} for local feature extraction and Transformers for global feature extraction, capturing both fine-grained details and broad contextual information, essential for generalizing across domains. Importantly, introducing prompts improves the performance of baseline methods. For instance, VPT is 3.2\% higher than ViT on average. The style prompt in StepSPT further demonstrates the significant impact of adaptive prompts on cross-domain downstream tasks.

\begin{table*}[!t]
  \caption{Comparison between SOTA methods and StepSPT in the 5-way 1-shot and 5-way 5-shot tasks. \textbf{SF} means if the method is source-free. \textbf{Backbone} indicates the utilized backbone in the method. \textbf{Freeze} exhibits if the backbone is frozen. `Y' means yes, `-' means no. The blue background cells from dark to light mean the best, second best, and third best performance.
  }
  \vspace{-2mm}
  \label{compareSOTAs}
  \footnotesize 
  \centering
  \setlength{\tabcolsep}{2.4mm}{
  \begin{tabular}{c|lccc|cccccc}
    \hline
    Task & \textbf{Methods} & \textbf{SF} & \textbf{Backbone} & \textbf{Freeze} & \textbf{CropDisease} & \textbf{EuroSAT} & \textbf{ISIC} & \textbf{ChestX} & \textbf{PatternNet} & \textbf{Avg}   \\
    \hline
     \multirow{18}*{\textbf{\rotatebox{270}{5-way 1-shot}}} & Finetuning~\cite{guo2020broader} & - & ResNet10 & - & 68.46$\pm$0.87 & 59.18$\pm$0.85 & 33.11$\pm$0.60 &  22.54$\pm$0.42 &  79.23$\pm$0.79 & 52.51    \\
    & LRP~\cite{sun2021explanation}      & - & ResNet10 & - & 59.23$\pm$0.50 & 54.99$\pm$0.50 & 30.94$\pm$0.30 & 22.11$\pm$0.20 &  - & -   \\
    & $\text{FDMixup}$~\cite{fu2021meta}     & - & ResNet10 & - & 66.23$\pm$1.03 & 62.97$\pm$1.01 & 32.48$\pm$0.64 & 22.26$\pm$0.45  & - & -    \\
     & FWT~\cite{tseng2020cross}      & - & ResNet10 & - & 66.36$\pm$1.04 & 62.36$\pm$1.05 & 31.58$\pm$0.67 & 22.04$\pm$0.44  &  - & -   \\
    & $\text{STARTUP}$~\cite{phoo2020self}     & - & ResNet10 & - & 75.93$\pm$0.80 & 63.88$\pm$0.84 & 32.66$\pm$0.60 & 23.09$\pm$0.43  &  - & -      \\
    & LDP-Net~\cite{zhou2023revisiting} & - & ResNet10 & - &  69.64 &  65.11 &  33.97 & 23.01  & - & -   \\
    & $\text{DDA}$~\cite{islam2021dynamic} & - & ResNet10 & - &  82.14$\pm$0.78 &  73.14$\pm$0.84 & 34.66$\pm$0.58 &  23.38$\pm$0.43   &  - & -      \\
    & \text{StyleAdv}~\cite{fu2023styleadv}  & - & ResNet10 & - & 80.69$\pm$0.28 & 72.92$\pm$0.75 & 35.76$\pm$0.52 & 22.64$\pm$0.35  &  - & -      \\
    & KT~\cite{li2023knowledge} & - & ResNet10 & - & 73.10$\pm$0.87 & 66.43$\pm$0.93 & 34.06$\pm$0.77 & 22.68$\pm$0.60 &  - & -      \\
    & \text{ReFine}~\cite{oh2022refine} & - & ResNet10 & - & 68.93$\pm$0.84 & 64.14$\pm$0.82 & 35.30$\pm$0.59 & 22.48$\pm$0.41  &  - &  -   \\
    & \text{VDB}~\cite{yazdanpanah2022visual}      & - & ResNet10 & Y & 71.98$\pm$0.82 & 63.60$\pm$0.87 & 35.32$\pm$0.65 &  22.99$\pm$0.44  &  - & -     \\
    & \text{RDC}~\cite{li2022ranking}  & - & ResNet10 & - & 86.33$\pm$0.50 &   71.57$\pm$0.50 &  35.84$\pm$0.40 & 22.27$\pm$0.20  & - & -     \\
    & $\text{CLDFD}$~\cite{zheng2023cross} & - & ResNet10 & - &  90.48$\pm$0.72 &  \cellcolor{blue!40}\textbf{82.52$\pm$0.76} &    \cellcolor{blue!40}\textbf{39.70$\pm$0.69} & 22.39$\pm$0.44 & \cellcolor{blue!10}94.94$\pm$0.61 & \cellcolor{blue!40}\textbf{66.01}      \\
    & DARA~\cite{zhao2023dual} & - & ResNet10 & - & 80.74$\pm$0.76 & 67.42$\pm$0.80 &  36.42$\pm$0.64 & 22.92$\pm$0.40 & 94.44$\pm$0.72 & 60.39      \\
    & IM-DCL~\cite{xu2024enhancing}     & Y & ResNet10 & - &   84.37$\pm$0.99 &  \cellcolor{blue!25}77.14$\pm$0.71 &  \cellcolor{blue!25}38.13$\pm$0.57 &   \cellcolor{blue!40}\textbf{23.98$\pm$0.79}  &  \cellcolor{blue!25}96.32$\pm$0.63 & \cellcolor{blue!10}63.99     \\  

    \cline{2-11} 
    & DARA~\cite{zhao2023dual} & - & ConvNeXt & Y & \cellcolor{blue!10}91.80$\pm$0.85 & 69.54$\pm$0.62 & 34.99$\pm$0.53 & 22.66$\pm$0.75 & 93.49$\pm$0.65 & 62.50      \\
    & IM-DCL~\cite{xu2024enhancing}     & Y & ConvNeXt & Y &  \cellcolor{blue!25}95.13$\pm$0.95 & 71.27$\pm$0.42 & 36.53$\pm$0.93 & \cellcolor{blue!10}23.53$\pm$0.26 & 93.47$\pm$0.57 & 63.99     \\
    
    & StepSPT (Ours) & Y & ConvNeXt & Y & \cellcolor{blue!40}\textbf{95.39$\pm$0.22} & \cellcolor{blue!10}73.83$\pm$0.80 & \cellcolor{blue!10}37.16$\pm$0.68 & \cellcolor{blue!25}23.73$\pm$0.98 & \cellcolor{blue!40}\textbf{96.44$\pm$0.42} & \cellcolor{blue!25}65.31     \\
    \hline
    \hline

    \multirow{19}*{\textbf{\rotatebox{270}{5-way 5-shot}}} & Finetuning~\cite{guo2020broader} & - & ResNet10 & - & 89.25$\pm$0.51 & 79.08$\pm$0.61 & 48.11$\pm$0.64 & 25.97$\pm$0.41 & 95.32$\pm$0.32 & 67.55    \\
    & LRP~\cite{sun2021explanation}      & - & ResNet10 & - & 86.15$\pm$0.40 & 77.14$\pm$0.40 & 44.14$\pm$0.40 & 24.53$\pm$0.30  &  - & -   \\
    & $\text{FDMixup}$~\cite{fu2021meta}     & - & ResNet10 & - & 87.27$\pm$0.69 & 80.48$\pm$0.79 & 44.28$\pm$0.66 & 24.52$\pm$0.44   &  - & -    \\
    & FWT~\cite{tseng2020cross}      & - & ResNet10 & - & 87.11$\pm$0.67 & 83.01$\pm$0.79 & 43.17$\pm$0.70 & 25.18$\pm$0.45  &  - & -   \\
    & \text{Confess}~\cite{das2022confess} & - & ResNet10 & - & 88.88$\pm$0.51 & 84.65$\pm$0.38 & 48.85$\pm$0.29 & 27.09$\pm$0.24   &  - & -    \\
    & $\text{STARTUP}$~\cite{phoo2020self}     & - & ResNet10 & - & 93.02$\pm$0.45 & 82.29$\pm$0.60 & 47.22$\pm$0.61 & 26.94$\pm$0.44  & - & -      \\
    & LDP-Net~\cite{zhou2023revisiting} & - & ResNet10 & - &  91.89 &  84.05 &  48.44 &  26.88 & - & -   \\
    & $\text{DDA}$~\cite{islam2021dynamic} & - & ResNet10 & - & 95.54$\pm$0.38 & 89.07$\pm$0.47 & 49.36$\pm$0.59 &  \cellcolor{blue!25}28.31$\pm$0.46  &  - & -      \\
    & \text{StyleAdv}~\cite{fu2023styleadv}  & - & ResNet10 & - &  96.51$\pm$0.28 &  \cellcolor{blue!25}91.64$\pm$0.43 &  53.05$\pm$0.54 & 26.24$\pm$0.35  &  - & -      \\
    & KT~\cite{li2023knowledge} & - & ResNet10 & - & 89.53$\pm$0.58 & 82.53$\pm$0.66 & 46.37$\pm$0.77 &  26.79$\pm$0.61  & - & -      \\
    & \text{ReFine}~\cite{oh2022refine} & - & ResNet10 & - & 90.75$\pm$0.49 & 82.36$\pm$0.57 &  51.68$\pm$0.63 & 26.76$\pm$0.42  &  - & -    \\
    & \text{VDB}~\cite{yazdanpanah2022visual}      & - & ResNet10 & Y & 90.77$\pm$0.49 & 82.06$\pm$0.63 & 48.72$\pm$0.65 & 26.62$\pm$0.45  &  - & -     \\
    & \text{RDC}~\cite{li2022ranking}  & - & ResNet10 & - &  93.55$\pm$0.30 &  84.67$\pm$0.30 & 49.06$\pm$0.30 & 25.48$\pm$0.20  & - & -     \\
    & $\text{CLDFD}$~\cite{zheng2023cross} & - & ResNet10 & - &  \cellcolor{blue!10}96.58$\pm$0.39 &  \cellcolor{blue!40}\textbf{92.89$\pm$0.34} & 52.29$\pm$0.62 & 25.98$\pm$0.43  & 99.00$\pm$0.28 & 73.35     \\
    & DARA~\cite{zhao2023dual} & - & ResNet10 & - &  95.32$\pm$0.34 &  85.84$\pm$0.54 & \cellcolor{blue!40}\textbf{56.28$\pm$0.66} &  \cellcolor{blue!10}27.54$\pm$0.42  & \cellcolor{blue!40}\textbf{99.26$\pm$0.22} & \cellcolor{blue!10}72.85      \\
    & IM-DCL~\cite{xu2024enhancing}     & Y & ResNet10 & - &  \textbf{95.73$\pm$0.38} &  \textbf{89.47$\pm$0.42} &  52.74$\pm$0.69 &  \cellcolor{blue!40}\textbf{28.93$\pm$0.41}  & \cellcolor{blue!10}99.00$\pm$0.85 & \cellcolor{blue!25}73.17     \\ 
    \cline{2-11}
    & DARA~\cite{zhao2023dual} & - & ConvNeXt & Y & 95.65$\pm$0.25 & 82.12$\pm$0.26 & 50.80$\pm$0.86 & 26.37$\pm$0.51 & 98.62$\pm$0.37 & 70.71      \\
    & IM-DCL~\cite{xu2024enhancing}     & Y & ConvNeXt & Y & \cellcolor{blue!40}\textbf{97.33$\pm$0.33} & 84.53$\pm$0.90 & \cellcolor{blue!25}54.33$\pm$0.37 & 26.40$\pm$0.98 & 99.00$\pm$0.42 & 72.32     \\ 
    
    & StepSPT (Ours) & Y & ConvNeXt & Y & \cellcolor{blue!25}97.11$\pm$0.60 & \cellcolor{blue!10}91.07$\pm$0.73 & \cellcolor{blue!10}53.99$\pm$0.44 & 27.11$\pm$0.75 & \cellcolor{blue!25}99.24$\pm$0.32 & \cellcolor{blue!40}\textbf{73.79}     \\
    \hline
  \end{tabular}
  \vspace{-5mm}
  }
\end{table*}

Similar to the 5-way 1-shot task, the average performance ranking of different backbones on the 5-shot task is: ConvNeXt (72.17\%) $\ge$ CLIP (70.29\%) $\ge$ ViT (67.89\%) $\ge$ ResNet10 (66.86\%). However, in the 5-way 5-shot task, CNN models no longer consistently outperform Transformer-based models on distant domains. For instance, ViT (47.66\%) and CLIP (48.67\%) perform better than ResNet10 (45.87\%) on ISIC. This outcome can be attributed to the ability of Transformer models to better leverage larger data volumes. As a result, in 5-way 5-shot tasks, with increased data volumes, the strengths of Transformer-based models become more evident. Furthermore, except for CoOp and CoCoOp, all prompt-based algorithms show performance improvements over their corresponding backbone baselines. Moreover, the proposed StepSPT outperforms all baselines and existing prompt-based methods on all datasets. StepSPT with the ConvNeXt backbone achieves the highest average performance (73.79\%) in the 5-way 5-shot task, indicating that ConvNeXt offers significant advantages for CDFSL tasks over other backbones like ResNet10 and ViT. Therefore, the combined capabilities of ConvNeXt make it the optimal choice for CDFSL tasks.


\subsection{Compare with State-of-the-art}
\label{sotas}
This section primarily compares StepSPT with existing state-of-the-art methods (SOTAs), as shown in Table~\ref{compareSOTAs}. However, ensuring fairness in these comparisons presents challenges. Specifically, most existing SOTAs are validated on ResNet10, require access to the source domain data, and employ specific training strategies to enhance cross-domain transferability. Moreover, these methods fine-tune the model during the adaptation stage. In contrast, the method proposed in this paper follows a source-free setting, which does not require access to source domain data or the use of specific training strategies. Additionally, it keeps the backbone parameters fixed during adaptation, without any fine-tuning. Based on the experimental results in Section~\ref{prompt_methods}, we have chosen ConvNeXt as the backbone instead of ResNet10. These differences in settings make direct comparisons between StepSPT and existing SOTAs challenging under varying conditions.

In this scenario, enforcing uniform comparison conditions might obscure each method’s performance in its preferred configuration. To ensure fairness, we present each method's optimal results to evaluate their performance more effectively under ideal conditions. To further enhance transparency, we also include results for two SOTA approaches (IM-DCL and DARA) on ConvNeXt, offering a comprehensive comparison across backbones.

As shown in Table~\ref{compareSOTAs}, in the 5-way 1-shot task, the proposed StepSPT achieves a relatively lower average result (65.31\%) compared to current SOTAs. Notably, StepSPT performs best on near domain tasks, achieving the highest results of 95.39\% and 96.44\% on CropDisease and PatternNet, respectively. However, StepSPT achieves the third-best performance on EuroSAT (73.83\%) and ISIC (37.16\%). Specifically, the lower sample resolution in EuroSAT contributes to a third-best result for StepSPT. In contrast, the high-resolution samples in PatternNet enable the model to achieve the best performance. Additionally, unlike other datasets with clean backgrounds, ISIC images are particularly challenging due to background interference. For instance, in the 'melanocytic nevus' category, images containing hair may cause the model to misidentify hair as a pathological feature. The model’s attention on ISIC images is easily distracted by background elements (e.g., hair), resulting in performance degradation. This issue is further discussed in Section~\ref{problem}. Moreover, compared to ConvNeXt-based DARA and IM-DCL, StepSPT achieves the best performance.

Furthermore, in the 5-way 5-shot task, StepSPT achieves the highest average performance of 73.79\%. StepSPT maintains strong performance on near domain tasks, achieving 97.11\% on CropDisease and 99.24\% on PatternNet, indicating that its advantage in the near domain declines with increasing data volume. Similar to the 5-way 1-shot task, StepSPT also obtains the third-best results on distant domains due to the sample characteristics as previously discussed. On average, StepSPT (73.79\%) outperforms ConvNeXt-based DARA (70.71\%) and IM-DCL (72.32\%).

\subsection{Ablation Study}

\subsubsection{Performance of Style Prompt}
We evaluate the performance of the style prompt from three aspects. First, we demonstrate the advantages of the style prompt by comparing results with and without the style prompt. Next, we compare our proposed style prompt with other prompts, including CoOp and CoCoOp. Finally, we visualize the statistical differences between samples with and without the style prompt.

\textbf{\textit{w/} and \textit{w/o} Style Prompt.}
We present the results of $w/$ and $w/o$ prompts in Table~\ref{tab:style}. Compared to the Baseline' (60.95\% and 72.17\% average results on 5-way 1-shot and 5-way 5-shot tasks), the style prompt (\textit{without StepSPT'}) improves performance on all datasets and overall (64.11\% and 73.50\% on average), demonstrating the effectiveness of prompts in CDFSL tasks. The impact of the style prompt can also be evaluated by comparing \textit{w/o} Style' and StepSPT.' Notably, StepSPT' consistently outperforms \textit{w/o} Style' across all datasets. For example, it achieves 0.4\% higher on CropDisease, 0.6\% on EuroSAT, 1.75\% on ISIC, 0.84\% on ChestX, 0.42\% on PatternNet, and an overall improvement of 0.6\% in the 5-way 5-shot task. This comparison clearly demonstrates the positive role of the style prompt combined with the step-wise distribution alignment strategy.
\begin{table}[!t]
  \caption{Performance of style prompt and step-wise distribution alignment strategy. 
  }
  \vspace{-3mm}
  \label{tab:style}
  \scriptsize 
  \centering
  \setlength{\tabcolsep}{0.7mm}{
  \begin{tabular}{c|l|cccccc}
    \hline
     \textbf{K} & \textbf{Strategy} & \textbf{CropDisease} & \textbf{EuroSAT} & \textbf{ISIC} & \textbf{ChestX} & \textbf{PatternNet} & \textbf{Average}   \\
    \hline
    \multirow{4}*{1} & Baseline & 91.35$\pm$0.64 & 66.25$\pm$0.77 & 33.13$\pm$0.53 & 22.11$\pm$0.55 & 91.93$\pm$0.25 & 60.95   \\
    & \textit{w/o} Step & 94.95$\pm$0.32 & 73.24$\pm$0.85 & 35.37$\pm$0.25 & 22.75$\pm$0.19 & 94.59$\pm$0.41 & 64.11   \\
    & \textit{w/o} Style & 93.38$\pm$0.50 & 73.11$\pm$0.34 & 36.67$\pm$0.86 & 22.76$\pm$0.12 & 93.96$\pm$0.81 & 63.98   \\
    & StepSPT & \textbf{95.39$\pm$0.22} & \textbf{73.83$\pm$0.80} & \textbf{37.16$\pm$0.68} & \textbf{23.73$\pm$0.98} & \textbf{96.44$\pm$0.42} & \textbf{65.31}     \\
    \hline
    \multirow{4}*{5} & Baseline & 95.15$\pm$0.29 & 88.11$\pm$0.90 & 52.51$\pm$0.48 & 26.57$\pm$0.21 & 98.52$\pm$0.36 & 72.17  \\
    & \textit{w/o} Step & \textbf{98.49$\pm$0.56} & 90.47$\pm$0.43 & 52.88$\pm$0.53 & 26.61$\pm$0.81 & 99.12$\pm$0.83 & 73.50   \\
    & \textit{w/o} Style & 96.71$\pm$0.85 & 90.47$\pm$0.57 & 52.24$\pm$0.83 & 26.27$\pm$0.93 & 98.82$\pm$0.34 & 73.10   \\  
    & StepSPT & 97.11$\pm$0.60 & \textbf{91.07$\pm$0.73} & \textbf{53.99$\pm$0.44} & \textbf{27.11$\pm$0.75} & \textbf{99.24$\pm$0.32} & \textbf{73.70}    \\
    \hline
  \end{tabular}
  }
\end{table}

\begin{figure}[!t]
  \centering
  \includegraphics[width=\linewidth]{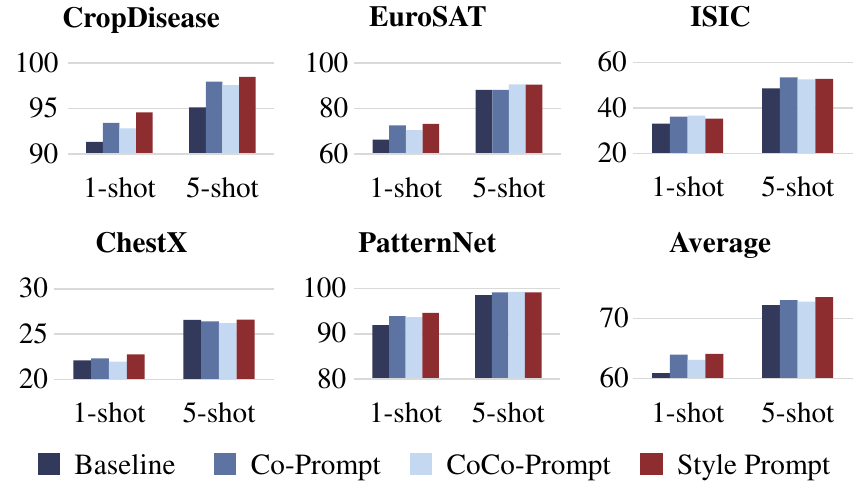}
  \vspace{-3mm}
  \caption{Comparison of style prompt and the current prompts in 5-way 1-shot and 5-way 5-shot tasks.}
  \vspace{-2mm}
  \label{fig:style}
\end{figure}

\textbf{Comparison between style prompt and other prompts.}
As shown in Figure~\ref{fig:style}, compared to Co-Prompt and CoCo-Prompt (where the style prompt in StepSPT is replaced by text prompts from CoOp and CoCoOp), the proposed style prompt achieves the best average performance. However, Co-Prompt and CoCo-Prompt achieve higher results than the style prompt on ISIC, indicating that, compared to visual prompts, text prompts can better alleviate performance degradation caused by background interference.

\begin{figure}[!t]
  \centering
  \includegraphics[width=\linewidth]{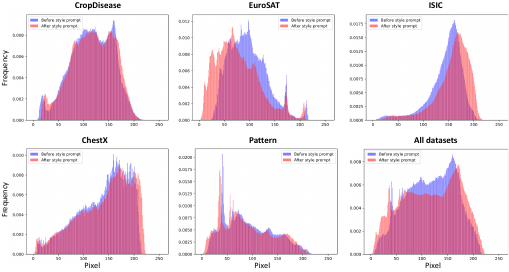}
  \vspace{-5mm}
  \caption{Histogram differences between data before and after style prompt tuning. Blue part means before style prompt, and red part indicates after style prompt.}
  \vspace{-2mm}
  \label{fig:statistic}
\end{figure}

\textbf{Style prompt visualization.}
This paper modifies image styles by adjusting the statistical characteristics in normalization. Current style transfer methods~\cite{huang2017arbitrary} use a nonlinear decoder to amplify the style effect, which introduces noise~\cite{gatys2016image}. Thus, we use histograms to show statistical characteristics rather than directly visualizing the style transfer. This approach allows us to observe the impact of style prompts on image statistics more directly, without the noise introduced by the nonlinear network. Figure~\ref{fig:statistic} shows the statistical differences in samples before and after applying the style prompt, which shows that the statistical characteristics significantly change after applying the style prompt.

\subsubsection{Performance of Step-wise Distribution Alignment}
We evaluate the proposed step-wise distribution alignment strategy from the following aspects. First, we demonstrate the performance of this strategy. Then, we choose the hyper-parameters related to this strategy, including step size, credible group, and loss function.

\textbf{\textit{w/} and \textit{w/o} Step-wise Distribution Alignment.}
As illustrated in Table~\ref{tab:style}, `\textit{w/o} Style' indicates StepSPT without the style prompt, \ie using only the step-wise distribution alignment strategy. Compared to `Baseline', `\textit{w/o} Style' significantly improves performance. For example, the average results of `\textit{w/o} Style' are 3.03\% (5-way 1-shot) and 0.93\% (5-way 5-shot) higher than `Baseline'. This indicates that the strategy is more effective in 5-way 1-shot, particularly in distant domain tasks such as ISIC and ChestX. In 5-way 1-shot task, `\textit{w/o} Style' is 3.54\% and 0.65\% higher than `Baseline' on ISIC and ChestX, respectively. However, in the 5-way 5-shot task, the strategy does not have a positive effect on these two datasets. Moreover, comparing `\textit{w/o} Step' and `StepSPT', combining this strategy with the style prompt further enhances performance. This suggests that, although the strategy decreases performance in distant domains in the 5-way 5-shot task, combining it with the style prompt yields a more positive effect (1.11\% higher on ISIC and 0.5\% higher on ChestX).

\textbf{Selection of Step Size.}
In step-wise distribution alignment strategy, the step size should be studied. Table~\ref{tab:step_size} presents the results of different step size. In 5-way 1-shot task, StepSPT obtains the best average result 65.31\% when the step size is 100. A step size of 100 achieves the best results on all datasets except PatternNet. Similar to 5-way 1-shot, the size of 100 also performs best on the 5-way 5-shot task, achieving 73.70\% optimal average result. Unlike the 5-way 1-shot task, a step size of 5 performs best only in distant domains (91.07\% for EuroSAT, 53.99\% for ISIC, and 27.11\% for ChestX) but not in near domain (third-optimal 97.11\% result for CropDisease). It is worth noting that a step size of 100 achieves an optimal result of 91.07\% on EuroSAT but the lowest result of 99.24\% on PatternNet. Despite this, we select the overall optimal step size of 100.
\begin{table}[!t]
  \caption{Performance of different step size \textit{E}. }
  \vspace{-2mm}
  \label{tab:step_size}
  \scriptsize 
  \centering
  \setlength{\tabcolsep}{1.0mm}{
  \begin{tabular}{c|c|cccccc}
    \hline
     \textbf{K} & \textbf{\textit{E}} & \textbf{CropDisease} & \textbf{EuroSAT} & \textbf{ISIC} & \textbf{ChestX} & \textbf{PatternNet} & \textbf{Average}   \\
    \hline
    \multirow{6}*{1} & 2 & 93.60$\pm$0.73 & 73.28$\pm$0.67 & 36.07$\pm$0.43 & 22.23$\pm$0.42 & 96.64$\pm$0.26 & 64.36   \\
    & 5 & 93.13$\pm$0.24 & 73.40$\pm$0.47 & 37.16$\pm$0.69 & 22.27$\pm$0.52 & 94.80$\pm$0.15 & 64.15   \\
    & 10 & 94.45$\pm$0.75 & 73.79$\pm$1.09 & 36.93$\pm$0.86 & 22.84$\pm$0.64 & \textbf{96.69$\pm$0.57} & 64.94   \\
    & 20 & 93.00$\pm$0.15 & 73.53$\pm$0.37 & 36.00$\pm$0.31 & 22.20$\pm$0.71 & 96.62$\pm$0.67 & 64.27   \\
    & 50 & 94.98$\pm$0.15 & 73.53$\pm$0.46 & 36.73$\pm$0.82 & 22.80$\pm$0.76 & 96.27$\pm$0.69 & 64.86   \\
    & 100 & \textbf{95.39$\pm$0.22} & \textbf{73.83$\pm$0.80} & \textbf{37.16$\pm$0.68} & \textbf{23.73$\pm$0.98} & 96.44$\pm$0.42 & \textbf{65.31}     \\
    \hline
    \multirow{6}*{5} & 2 & 96.64$\pm$0.50 & 89.00$\pm$0.19 & 50.93$\pm$0.17 & 25.80$\pm$0.75 & \textbf{99.53$\pm$0.38} & 72.38   \\
    & 5 & 96.40$\pm$0.15 & 90.06$\pm$0.67 & 51.73$\pm$0.28 & 25.37$\pm$0.25 & 99.26$\pm$0.67 & 72.56   \\
    & 10 & \textbf{98.67$\pm$0.76} & 89.63$\pm$0.30 & 52.05$\pm$0.51 & 26.80$\pm$0.33 & 99.40$\pm$0.28 & 73.31   \\
    & 20 & 98.62$\pm$0.89 & 89.99$\pm$0.17 & 51.89$\pm$0.80 & 25.60$\pm$0.53 & 99.33$\pm$0.47 & 73.09   \\
    & 50 & 97.10$\pm$0.69 & 90.20$\pm$0.71 & 52.20$\pm$0.62 & 27.07$\pm$0.94 & 99.33$\pm$0.79 & 73.18   \\  
    & 100 & 97.11$\pm$0.60 & \textbf{91.07$\pm$0.73} & \textbf{53.99$\pm$0.44} & \textbf{27.11$\pm$0.75} & 99.24$\pm$0.32 & \textbf{73.70}    \\
    \hline
  \end{tabular}
  \vspace{-5mm}
  }
\end{table}

\textbf{Credible Group.}
As written in the paper, we obtain the credible group $\mathcal{G}$ in two ways: entropy-based ranking $\mathcal{G}_{en}$, and prototype-based ranking $\mathcal{G}_{pro}$. In Table~\ref{tab:credible}, we demonstrate the effectiveness of each group. Compared to using only $\mathcal{G}_{en}$ or $\mathcal{G}_{pro}$ separately, $\mathcal{G}$ performs best on all 5 datasets (0.56\% higher on CropDisease, 0.02\% higher on EuroSAT, 3.07\% higher on ISIC, 0.22\% higher on ChestX, 0.48\% higher on PatternNet, and 0.97\% higher on average) in 5-way 1-shot task. However, in the 5-way 5-shot task, $\mathcal{G}$ achieves sub-optimal results on CropDisease, EuroSAT, and PatternNet. Specifically, $\mathcal{G}_{en}$ performs best on CropDisease (98.29\%), while $\mathcal{G}_{pro}$ achieves the best performance on EuroSAT (91.22\%) and PatternNet (99.73\%). Therefore, $\mathcal{G}_{en}$ can make up for the performance shortcomings on CropDisease by introducing a small amount of labeled data, while $\mathcal{G}_{pro}$’s performance shortcomings in remote sensing also can be compensated by adding more labeled data.
\begin{table}[!t]
  \caption{Performance of different credible group with a ConvNeXt backbone.}
  \vspace{-2mm}
  \label{tab:credible}
  \scriptsize 
  \centering
  \setlength{\tabcolsep}{0.8mm}{
  \begin{tabular}{c|l|cccccc}
    \hline
     \textbf{K} & \textbf{Group} & \textbf{CropDisease} & \textbf{EuroSAT} & \textbf{ISIC} & \textbf{ChestX} & \textbf{PatternNet} & \textbf{Average}   \\
    \hline
    \multirow{3}*{1} & $\mathcal{G}_{en}$ & 94.27$\pm$0.82 & 72.00$\pm$0.26 & 34.09$\pm$0.34 & 23.51$\pm$0.77 & 94.80$\pm$0.35 & 63.73   \\
    & $\mathcal{G}_{pro}$ & 94.83$\pm$0.41 & 73.81$\pm$0.65 & 33.82$\pm$0.28 & 23.27$\pm$1.10 & 95.96$\pm$0.81 & 64.34   \\
    & $\mathcal{G}$ & \textbf{95.39$\pm$0.22} & \textbf{73.83$\pm$0.80} & \textbf{37.16$\pm$0.68} & \textbf{23.73$\pm$0.98} & \textbf{96.44$\pm$0.42} & \textbf{65.31}     \\
    \hline
    \multirow{3}*{5} & $\mathcal{G}_{en}$ & \textbf{98.29$\pm$0.53} & 89.82$\pm$0.44 & 52.96$\pm$0.34 & 26.62$\pm$0.66 & 99.47$\pm$0.32 & 73.43  \\
    & $\mathcal{G}_{pro}$ & 96.00$\pm$0.87 & \textbf{91.22$\pm$0.53} & 51.24$\pm$0.88 & 25.29$\pm$0.45 & \textbf{99.73$\pm$0.29} & 72.70   \\  
    & $\mathcal{G}$ & 97.11$\pm$0.60 & 91.07$\pm$0.73 & \textbf{53.99$\pm$0.44} & \textbf{27.11$\pm$0.75} & 99.24$\pm$0.32 & \textbf{73.70}    \\
    \hline
  \end{tabular}
  }
\end{table}

\begin{figure}[!t]
  \centering
  \includegraphics[width=\linewidth]{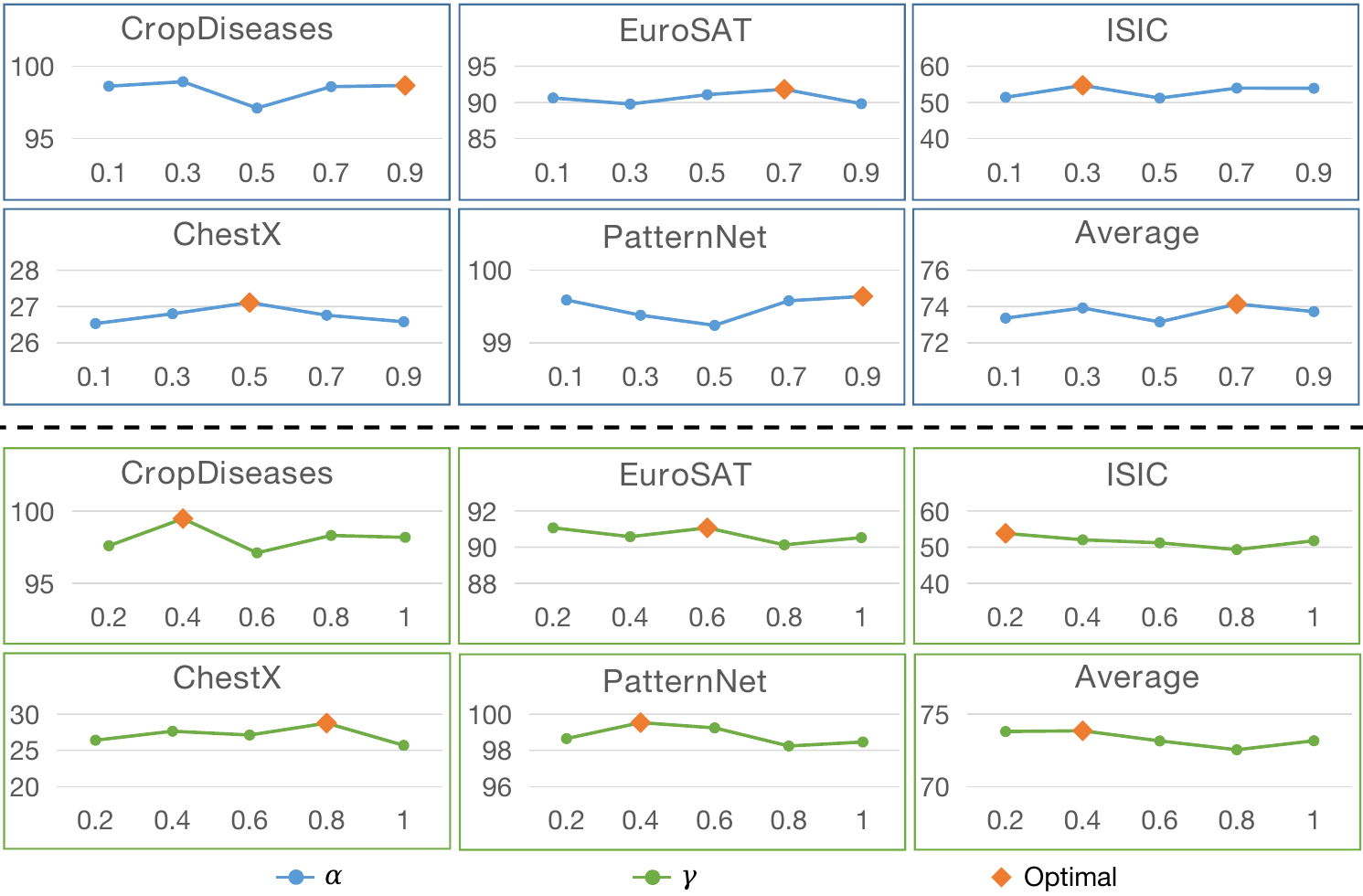}
  \vspace{-5mm}
  \caption{Results of StepSPT with different $\alpha$ and $\gamma$ on 5-way 5-shot task. The upper part means the performance of $\alpha$ on different datasets, and the down part is the selection of $\gamma$ on all datasets.}
  \vspace{-2mm}
  \label{fig:alpha}
\end{figure}

Additionally, the number of selected samples in $\mathcal{G}_{en}$ and $\mathcal{G}_{pro}$ depends on $\alpha$ and $\gamma$. Therefore, Figure~\ref{fig:alpha} illustrates the selection of $\alpha$ and $\gamma$. For $\alpha$, we choose from values [0.1, 0.3, 0.5, 0.7, 0.9]. The optimal values of $\alpha$ for all datasets (CropDisease to PatternNet) are 0.9, 0.7, 0.3, 0.5, and 0.9. We observe that $\alpha$ tends to the large value for addressing the near domains (0.9 for CropDisease, 0.7 for EuroSAT, and 0.9 for PatternNet), and the small value for distant domains (0.3 for ISIC and 0.5 for ChestX). This non-uniformity across domains suggests that sub-optimal values (0.7, 0.5, 0.7, 0.3, 0.7) offer a potential for setting $\alpha$ at 0.7. Moreover, the average optimal results of all datasets is 0.7. Therefore, the best selection of $\alpha$ is 0.7. Similarly for $\gamma$, we choose from [0.2, 0.4, 0.6, 0.8, 1.0], where the best selections from CropDisease to PatternNet are: 0.4, 0.6, 0.2, 0.8, 0.4. Following the selection of $\alpha$, we also analyze sub-optimal values of $\gamma$ across datasets: 0.8, 0.4, 0.4, 0.4, 0.6. And the best average result of $\gamma$ is 0.4, which guides us to set the optimal result of $\gamma$ to 0.4. These configurations (0.7 for $\alpha$ and 0.4 for $\gamma$) suggest that selecting more top-k samples with lower entropy enhances $\mathcal{G}$'s reliability, while decreasing $\gamma$ improves performance, indicating a preference for fewer \textit{topk} prototype samples.

\begin{table}[!t]
  \caption{Comparison of KL and MI loss effectiveness in 5-way 5-shot task.}
  \vspace{-3mm}
  \label{klmi}
  \scriptsize 
  \centering
  \setlength{\tabcolsep}{1.0mm}{
  \begin{tabular}{l|cccccc}
    \hline
     \textbf{Methods} & \textbf{CropDisease} & \textbf{EuroSAT} & \textbf{ISIC} & \textbf{ChestX} & \textbf{PatternNet} & \textbf{Avg}   \\
    \hline
     Baseline & 95.15$\pm$0.29 & 88.11$\pm$0.90 & 52.51$\pm$0.48 & 26.57$\pm$0.21 & 98.52$\pm$0.36 & 72.17    \\ 
    \textit{w/o} MI & 96.49$\pm$0.69 & 90.64$\pm$0.32 & 51.38$\pm$0.79 & 26.18$\pm$0.80 & \textbf{99.69$\pm$0.32} & 72.88      \\
     \textit{w/o} KL & 96.31$\pm$0.33 & 90.08$\pm$0.61 & 53.82$\pm$0.53 &  26.40$\pm$0.62 & 99.07$\pm$0.70 & 73.34    \\
    StepSPT & \textbf{97.11$\pm$0.60} & \textbf{91.07$\pm$0.73} & \textbf{53.99$\pm$0.44} & \textbf{27.11$\pm$0.75} & 99.24$\pm$0.32 & \textbf{73.79}      \\
    \hline
  \end{tabular}
  \vspace{-5mm}
  }
\end{table}

\textbf{Loss Function $L_{MI}$ and $L_{KL}$.}
Table~\ref{klmi} highlights the effectiveness of $KL$ and $MI$ loss within the step-wise distribution alignment strategy. As shown in the table, \textit{w/o} MI' outperforms Baseline' in near domains such as CropDisease (96.49\% vs. 95.15\%), EuroSAT (90.64\% vs. 88.11\%), and PatternNet (99.69\% vs. 98.52\%), but underperforms in distant domains like ISIC (51.38\% vs. 52.51\%) and ChestX (26.18\% vs. 26.57\%). An analysis suggests that the KL loss considers only sample diversity and ensures that samples are evenly classified into each category. However, it ignores the constraints of distance between data pairs, leading to reduced performance in distant domains. While In contrast, `\textit{w/o} KL' outperforms `Baseline' across nearly all datasets (96.31\% vs. 95.15\% for CropDisease, 90.08\% vs. 88.11\% for EuroSAT, 53.82\% vs. 52.51\% for ISIC, and 99.07\% vs. 98.52\% for PatternNet). This improvement is attributed to the MI loss, which reduces the distance between data pairs, resulting in the model being able to classify the sample deterministically even in the distant domain. Additionally, the average performance of `\textit{w/o} KL' surpasses `\textit{w/o} MI' (72.88\%), suggesting that MI loss is more crucial than KL loss. Moreover, combining MI and KL losses in StepSPT' further enhances classification performance, which emphasizes the importance of reducing distances of data pairs and maintaining diversity.

\begin{table*}[!t]
  \caption{Performance of Label Propagation (LP) on all 5 datasets in 5-way $K$-shot task.}
  \vspace{-3mm}
  \label{lplp}
  \scriptsize 
  \centering
  \setlength{\tabcolsep}{4.5mm}{
  \begin{tabular}{c|l|l|cccccc}
    \hline
     \textbf{K} & \textbf{Backbone} & \textbf{Methods} & \textbf{CropDisease} & \textbf{EuroSAT} & \textbf{ISIC} & \textbf{ChestX} & \textbf{PatternNet} & \textbf{Avg}   \\
    \hline
    \multirow{11}*{1} & \multirow{3}*{ViT} & VPT & 75.00$\pm$0.24 &  67.53$\pm$0.59 &  30.80$\pm$0.96  & 20.56$\pm$0.64  & 79.87$\pm$0.69  &  54.75   \\
    &  & StepSPT \textit{w/o} LP & 81.00$\pm$0.33  & 67.35$\pm$0.59  & 31.01$\pm$0.49  & 23.06$\pm$0.62  & 79.63$\pm$0.73  & 56.41    \\
    &  & StepSPT \textit{w/} LP & \cellcolor{blue!25}\textbf{82.56$\pm$0.27} & \cellcolor{blue!25}\textbf{68.47$\pm$0.22} & \cellcolor{blue!25}\textbf{32.03$\pm$0.42} & \cellcolor{blue!25}\textbf{23.55$\pm$1.06} & \cellcolor{blue!25}\textbf{80.84$\pm$0.31} & \cellcolor{blue!25}\textbf{57.49}    \\
    \cline{2-9}

    & \multirow{3}*{CLIP} & CoCoOp & 75.37$\pm$0.78  & \cellcolor{blue!25}\textbf{71.56$\pm$0.85}  & 30.07$\pm$0.36  & 20.23$\pm$0.84  & 90.55$\pm$0.31  & 57.56    \\
    &  & StepSPT \textit{w/o} LP & 82.69$\pm$0.64  & 70.25$\pm$0.31  & 31.88$\pm$0.83  & 22.62$\pm$0.28  & 92.47$\pm$0.57  & 59.98    \\
    &  & StepSPT \textit{w/} LP & \cellcolor{blue!25}\textbf{84.84$\pm$0.72} & 70.01$\pm$0.21 & \cellcolor{blue!25}\textbf{32.97$\pm$0.27} & \cellcolor{blue!25}\textbf{22.84$\pm$0.95} & \cellcolor{blue!25}\textbf{95.16$\pm$0.51} & \cellcolor{blue!25}\textbf{61.16}    \\
    \cline{2-9}

    & \multirow{5}*{ConvNeXt} & DARA & 91.80$\pm$0.85 &  69.54$\pm$0.62 & 34.99$\pm$0.53 & 22.66$\pm$0.75 & 93.49$\pm$0.65 & 62.50    \\
    &  & IM-DCL \textit{w/o} LP & 92.60$\pm$0.50  & 70.07$\pm$0.60  & 34.53$\pm$0.97  & 21.60$\pm$0.97  & 91.79$\pm$0.44 & 62.12    \\
    &  & StepSPT \textit{w/o} LP & 92.17$\pm$0.43 & 70.53$\pm$0.47 & 34.62$\pm$0.77 &  23.33$\pm$0.91 & 92.44$\pm$0.62 & 62.62   \\
    &  & IM-DCL \textit{w/} LP & 95.13$\pm$0.95 & 71.27$\pm$0.42 & 36.53$\pm$0.93 & 23.53$\pm$0.26 & 93.47$\pm$0.57 & 63.99    \\
    &  & StepSPT \textit{w/} LP & \cellcolor{blue!25}\textbf{95.39$\pm$0.22} & \cellcolor{blue!25}\textbf{73.83$\pm$0.80} & \cellcolor{blue!25}\textbf{37.16$\pm$0.66} & \cellcolor{blue!25}\textbf{23.73$\pm$0.98} & \cellcolor{blue!25}\textbf{96.44$\pm$0.42} & \cellcolor{blue!25}\textbf{65.31}      \\
    \cline{2-9}
    \hline
    \hline
    \multirow{11}*{5} & \multirow{3}*{ViT} & VPT & 95.20$\pm$0.99  & 78.05$\pm$0.39 &  50.00$\pm$0.44 &  26.63$\pm$1.03 &  93.24$\pm$0.74  &  68.62   \\
    &  & StepSPT \textit{w/o} LP & 95.39$\pm$0.49  & 82.51$\pm$0.24 &  49.95$\pm$0.36 &  25.31$\pm$0.84 &  93.88$\pm$0.66 &  69.41    \\
    &  & StepSPT \textit{w/} LP &\cellcolor{blue!25}\textbf{ 97.07$\pm$0.75} & \cellcolor{blue!25}\textbf{84.88$\pm$0.28} & \cellcolor{blue!25}\textbf{50.39$\pm$0.34} & \cellcolor{blue!25}\textbf{26.65$\pm$0.11} & \cellcolor{blue!25}\textbf{95.96$\pm$0.65} & \cellcolor{blue!25}\textbf{70.99}    \\
    \cline{2-9}

    & \multirow{3}*{CLIP} & CoCoOp & 90.20$\pm$0.34 &  86.07$\pm$0.62 &  46.99$\pm$0.71 &  26.15$\pm$0.83 &  98.71$\pm$0.40  & 69.62    \\
    &  & StepSPT \textit{w/o} LP & 93.53$\pm$0.37 &  87.52$\pm$0.29 &  49.98$\pm$0.14 &  26.29$\pm$0.59 &  98.64$\pm$0.81 &   71.19   \\
    &  & StepSPT \textit{w/} LP & \cellcolor{blue!25}\textbf{96.01$\pm$0.88} & \cellcolor{blue!25}\textbf{89.40$\pm$1.05} & \cellcolor{blue!25}\textbf{52.12$\pm$0.36} & \cellcolor{blue!25}\textbf{26.36$\pm$0.97} & \cellcolor{blue!25}\textbf{99.04$\pm$0.31} & \cellcolor{blue!25}\textbf{72.58}    \\
    \cline{2-9}

    & \multirow{5}*{ConvNeXt} & DARA & 95.65$\pm$0.25 &  82.12$\pm$0.26 &  50.80$\pm$0.86 &  26.37$\pm$0.51 &  98.62$\pm$0.37 &  70.71    \\
    &  & IM-DCL \textit{w/o} LP & 96.33$\pm$0.33 &  82.33$\pm$0.81 &  50.93$\pm$0.48 &  27.03$\pm$0.71 &  98.00$\pm$0.41 &  70.92    \\
    &  & StepSPT \textit{w/o} LP & 96.53$\pm$0.52 & 89.20$\pm$0.88 & 50.62$\pm$0.62 & 27.06$\pm$0.27 & 98.71$\pm$0.49 & 72.42    \\
    &  & IM-DCL \textit{w/} LP & \cellcolor{blue!25}\textbf{97.33$\pm$0.33} & 84.53$\pm$0.90 & \cellcolor{blue!25}\textbf{54.33$\pm$0.37} & 26.40$\pm$0.98 & 99.00$\pm$0.42 & 72.32    \\
    &  & StepSPT \textit{w/} LP & 97.11$\pm$0.60 & \cellcolor{blue!25}\textbf{91.07$\pm$0.73} & 53.99$\pm$0.44 & \cellcolor{blue!25}\textbf{27.11$\pm$0.75} & \cellcolor{blue!25}\textbf{99.24$\pm$0.32} & \cellcolor{blue!25}\textbf{73.79}      \\
    \hline
  \end{tabular}
  }
\end{table*}

\begin{figure}[!t]
  \centering
  \includegraphics[width=\linewidth]{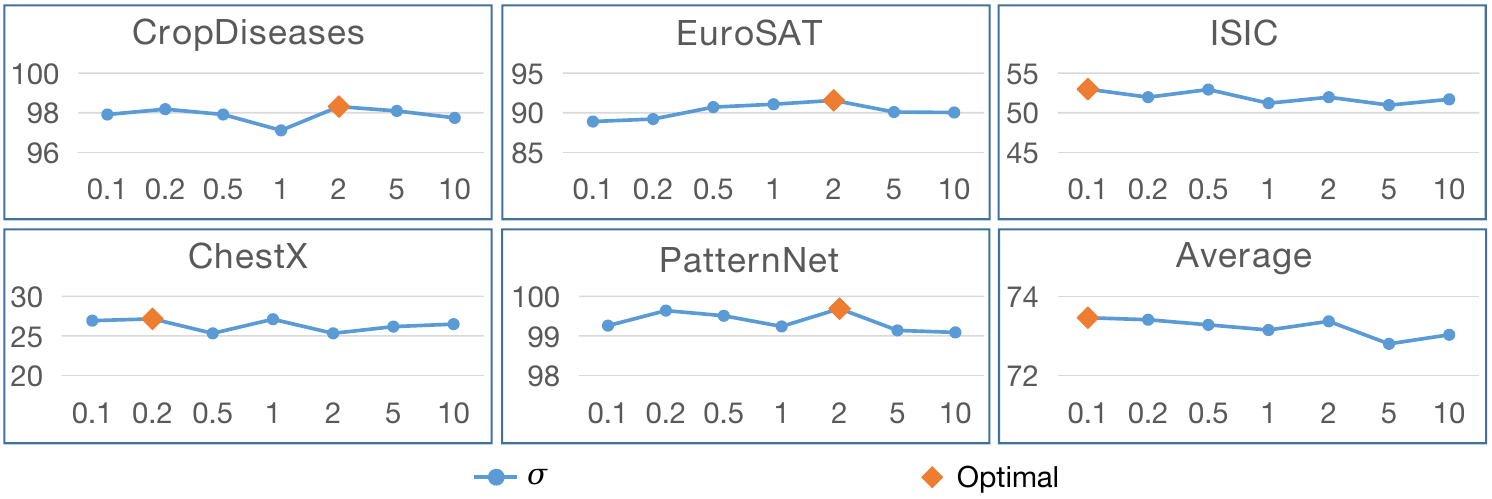}
  \vspace{-5mm}
  \caption{Results of StepSPT with different $\sigma$ on all the datasets in 5-way 5-shot task.}
  \vspace{-5mm}
  \label{fig:sigma}
\end{figure}

In addition, the hyper-parameter $\sigma$ is also important for $L_{ex}$, which controls the strength of the KL loss effect. Figure~\ref{fig:sigma} show the performance of different $\sigma$ in 5-way 5-shot task. The figure shows that the optimal $\sigma$ differs between near (CropDisease, EuroSAT, and PatternNet) and distant domains (ISIC and ChestX). Specifically, in near domains, the best performance is achieved (98.31\% for CropDisease, 91.56\% for EuroSAT, and 99.69\% for PatternNet) when $\sigma$ is set to a high value of 2. While in the distant domains, the optimal set of $\sigma$ is small like 0.1 or 0.2. For example, the best result on ISIC is 52.98\% when $\sigma$=0.1, and the optimal performance on ChestX is 27.16\% when $\sigma$ is 0.2. This suggests that KL loss plays a more limited role in distant domains, so a lower weight is preferred. This coincides with the conclusion obtained in Table~\ref{klmi}, where `\textit{w/o} KL' corresponds to $\sigma$ set to 0, and `\textit{w/o} MI' is similar to that $\sigma$ is set to an infinity constant. Figure~\ref{fig:sigma} further illustrates paying too much attention to KL loss will have a negative effect on distant domain tasks due to ignoring the uncertainty constraints of MI loss. Therefore, $\sigma$ is set to 2 for near domains and 0.1 for distant domains.

\subsection{Performance Analysis}
In this section, we focus on the analysis of StepSPT’s performance from three perspectives: (1) introduction of Label Propagation (LP), (2) generalization to the extra query set, and (3) T-SNE visualization of the proposed StepSPT.

\begin{figure}[!t]
  \centering
  \includegraphics[width=\linewidth]{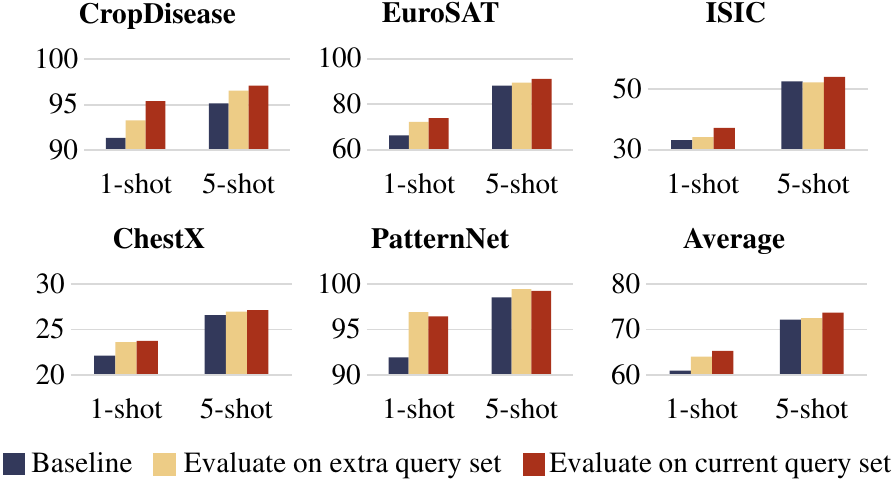}
  \vspace{-5mm}
  \caption{Evaluate on extra query set on all 5 datasets in both 5-way 1-shot and 5-way 5-shot tasks.}
  \vspace{-5mm}
  \label{fig:query}
\end{figure}

\subsubsection{Label Propagation (LP)}
To further demonstrate the performance of StepSPT, we compare `StepSPT \textit{w/o} LP' with the current SOTAs, such as VPT, CoCoOp, DARA, and IM-DCL, and `StepSPT \textit{w/} LP' on different backbones. The comparison between `StepSPT \textit{w/o} LP' and SOTAs demonstrates StepSPT’s superior performance, and `StepSPT \textit{w/} LP' can further demonstrate the effect of LP, as shown in Table~\ref{lplp}. For the ViT backbone, `VPT' outperforms `StepSPT \textit{w/o} LP' on EuroSAT (0.18\% higher) and PatternNet (0.24\% higher), indicating that ViT-based StepSPT has limited effectiveness in remote sensing. However, this limitation disappears in the 5-way 5-shot task, where `StepSPT \textit{w/o} LP' outperforms `VPT' on all datasets (4.46\% higher on EuroSAT and 0.64\% higher on PatternNet). Additionally, `StepSPT \textit{w/} LP' achieves the best results compared to `StepSPT \textit{w/o} LP' and `VPT'. For the CLIP backbone, we compare `StepSPT \textit{w/o} LP' and `StepSPT \textit{w/} LP' with `CoCoOp'. Since CLIP is based on ViT, similar to ViT backbone, `CoCoOp' similarly achieves the best performance on EuroSAT (71.56\%) in the 5-way 1-shot task, while `StepSPT \textit{w/o} LP' obtains 70.25\%. And same to ViT backbone, this phenomenon disappears as the amount of data increases, \ie 87.52\% of `StepSPT \textit{w/o} LP' is higher than 86.07\% of `CoCoOp' on EuroSAT in the 5-way 5-shot task. And `StepSPT \textit{w/} LP' is better than the other two, meaning the effect of LP. 

Moreover, for the ConvNeXt backbone, we first compare `StepSPT \textit{w/o} LP' with `DARA' and `IM-DCL \textit{w/o} LP' to demonstrate the superiority of StepSPT. Specifically, `StepSPT \textit{w/o} LP' outperforms `IM-DCL \textit{w/o} LP' and `DARA' on almost all datasets. For instance,`StepSPT \textit{w/o} LP' achieves average scores of 62.62\% and 72.42\% in the 5-way 1-shot and 5-way 5-shot tasks, respectively, compared to (62.12\%, 70.92\%) and (62.50\%, 70.71\%) for the other two. In addition, `StepSPT \textit{w/} LP' performs best on average (65.31\%, 73.79\%) in both 5-way 1-shot and 5-way 5-shot tasks compared to `DARA' (62.50\%, 70.71\%), `StepSPT \textit{w/o} LP' (62.62\%, 72.42\%) and `IM-DCL \textit{w/} LP' (63.99\%, 72.32\%). In general, we draw the following conclusions: (1) StepSPT still achieves the best results even without the introduction of LP, (2) the introduction of LP further improves the classification performance, and (3) StepSPT achieves the best performance among SOTAs that incorporate LP.

\subsubsection{Generalize to Extra Query Set}
Given that StepSPT introduces the query set into the external process, it is essential to evaluate its performance on the additional query set. The results of StepSPT on the additional query set are shown in Figure~\ref{fig:query}.
From Figure~\ref{fig:query} we observe that the extra query set results outperform the baseline, indicating the effectiveness of transductive learning in StepSPT. Additionally, the extra query set results are also close to those of the current query set, with the extra query set even outperforming the current query set on PatternNet, demonstrating StepSPT’s strong generalization ability. 
Moreover, this trend remains consistent on both 5-way 1-shot and 5-way 5-shot tasks, further indicating the robustness of transductive learning in the proposed method, \ie its positive effect remains stable as the data amount increases.


\begin{figure*}[!t]
  \centering
  \includegraphics[width=0.8\linewidth]{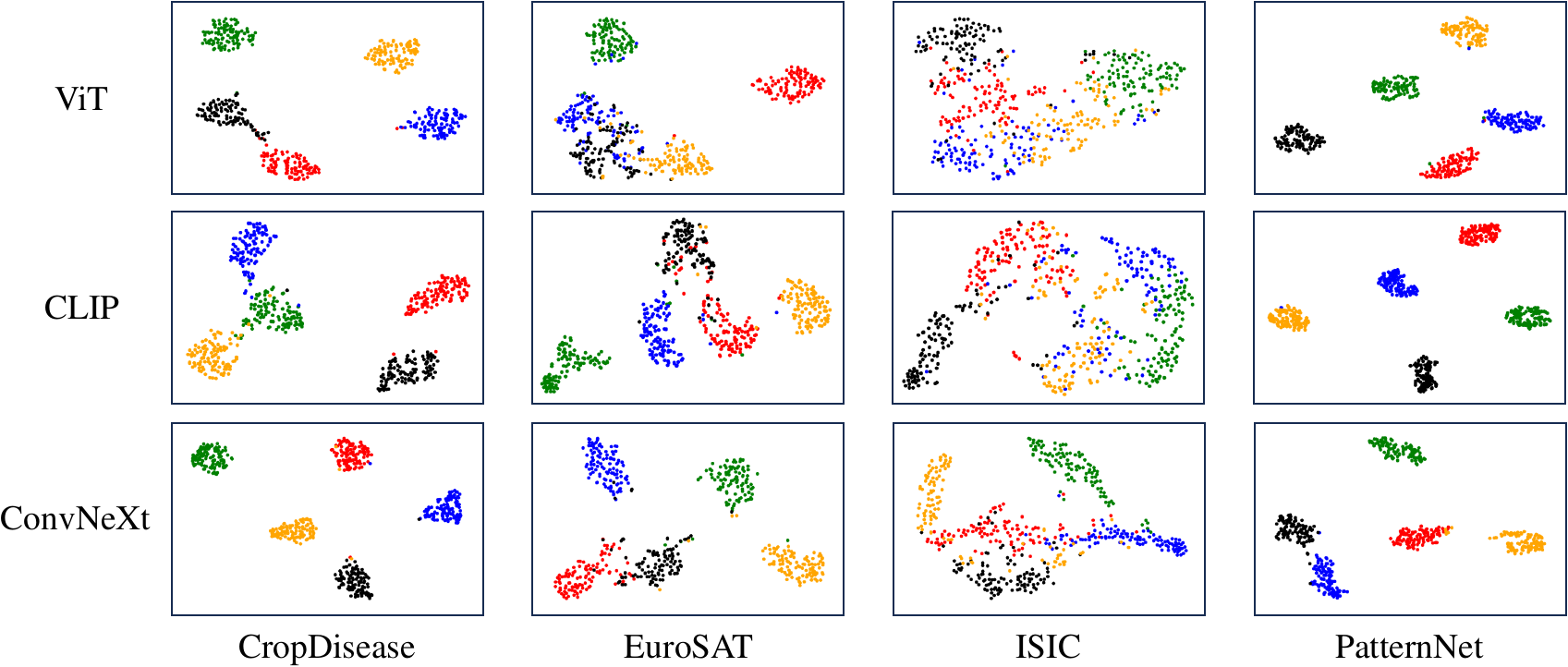}
  \vspace{-2mm}
  \caption{T-SNE visualization of StepSPT on different backbones: ViT, CLIP, and ConvNeXt.}
  \vspace{-4mm}
  \label{fig:tsne}
\end{figure*}

\begin{figure}[!t]
  \centering
  \includegraphics[width=0.9\linewidth]{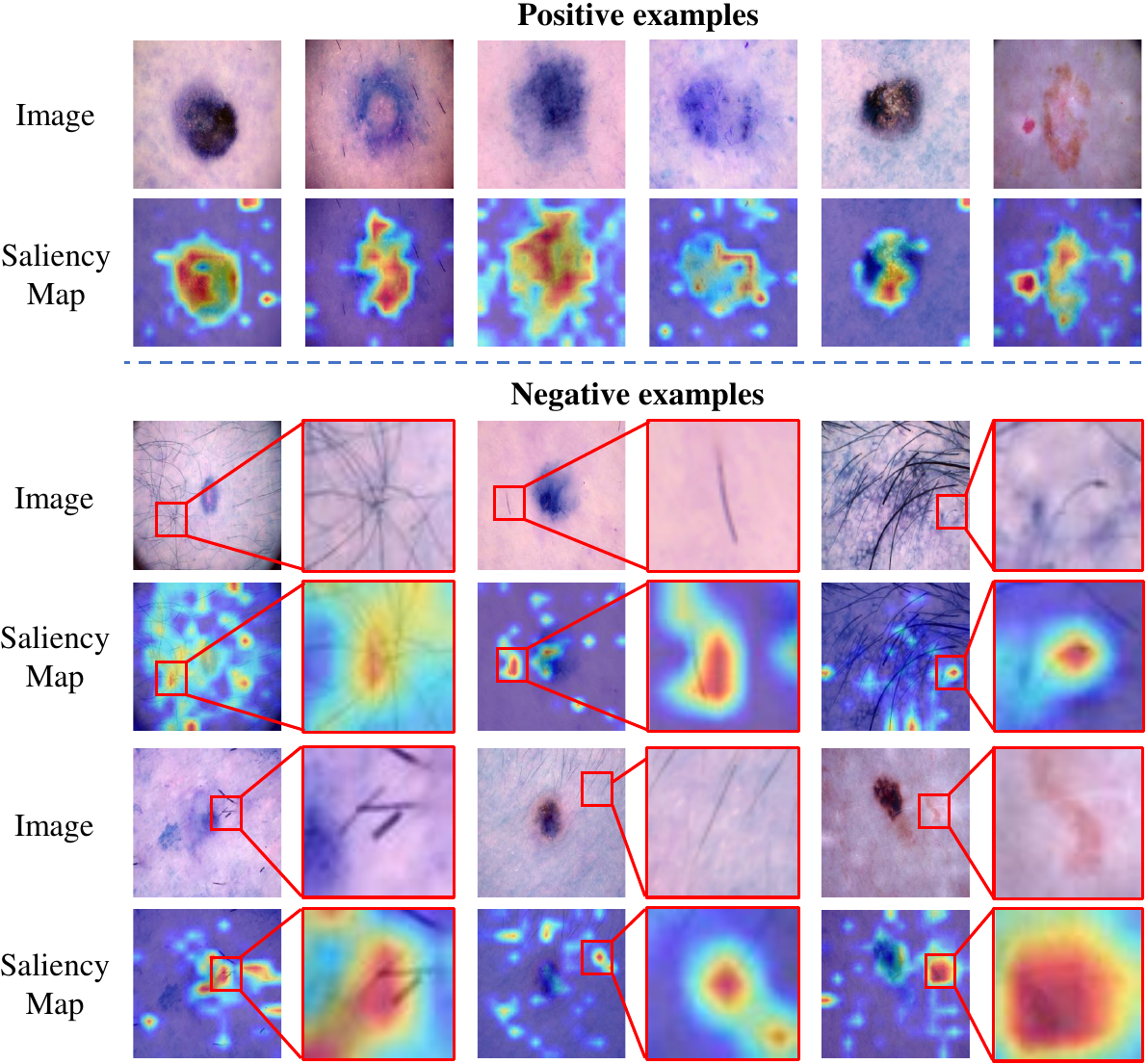}
  \vspace{-3mm}
  \caption{Positive (correct attention) and negative (incorrect attention) samples on ISIC dataset. The upper row shows the images, and the lower row is the corresponding saliency map to indicate the attention of model.}
  \vspace{-3mm}
  \label{fig:isic}
\end{figure}

\subsubsection{T-SNE Visualization}
\label{problem}
%
Due to the generally low performance on the 5-way 1-shot task, t-SNE~\cite{van2008visualizing} visualization cannot effectively show the characteristics of the proposed method, we present the visualization only for the 5-way 5-shot task, as shown in Figure~\ref{fig:tsne}. Additionally, we omit the visualization of ChestX due to its low performance. We observe that CropDisease and PatternNet (with high accuracy) show highly aggregated visual features. However, the separation boundaries of EuroSAT and ISIC (with low accuracy) are relatively vague, which means StepSPT does not have strict constraints on the certainty of features, especially on EuroSAT and ISIC. We use the following example of this phenomenon: a sample that belongs to the third category but gets a prediction result of [0.1, 0.31, 0.3, 0.2, 0.1] will be incorrectly classified to the second category. The high uncertainty of these features significantly affects classification accuracy. As discussed in Section~\ref{sotas}, this phenomenon is attributed to the low resolution of EuroSAT samples and the background interference in ISIC images.
Comparing the visualization of EuroSAT and PatternNet in Figure~\ref{fig:tsne} highlights the importance of resolution, where EuroSAT, compared to PatternNet, shows a tendency towards uncertain classification. 


\section{Limitation and Future Work}
Although StepSPT demonstrates promising results for SF-CDFSL, the method still has limitations that warrant further investigation. First, the model is sensitive to background interference, particularly in datasets such as ISIC where background artifacts (\eg hair) frequently appear. As discussed in Section~\ref{sotas}, unlike datasets with relatively clean backgrounds, ISIC images often contain distracting elements that can be spuriously learned as discriminative cues. For example, in the melanocytic nevus class, the presence of hair can mislead the model to treat it as a pathological feature. Because StepSPT performs step-wise distribution alignment without explicitly distinguishing foreground and background, background signals can enter the alignment and accumulate across steps. This not only distracts the model but also reduces classification accuracy. Figure~\ref{fig:isic} shows saliency maps~\cite{adebayo2018sanity}: in error cases, the model places its highest responses on background elements such as hair rather than on the lesion.

To understand why this happens, we examine the mechanism. The backbone responds strongly to hair because it presents sharp, high-contrast edges, whereas lesion cues are lower contrast and more diffuse. Because credible groups are chosen solely by confidence and consistency, and there is no explicit separation between foreground and background, early pseudo-labels that seem “confident” for hair-driven reasons are admitted, steering the stepwise alignment toward background textures. The large hair-induced activations dominate the updates, shifting class prototypes and decision margins toward background features, regardless of lesion size. Each step reinforces this shift, so later credible groups inherit the same bias. Since the current design does not explicitly separate foreground from background, background-driven activations are not down-weighted, which increases the chance that unreliable samples enter the credible group and propagate errors. Looking ahead, we will focus on mitigating background interference and improving the handling of uncertain predictions~\cite{liu2025causal}. The goal is to stabilize credible group selection and stepwise alignment so that early errors do not propagate and the model remains lesion-focused.

\section{Conclusion}
This paper introduces Step-wise Distribution-aligned Style Prompt Tuning (StepSPT) to address the source-free cross-domain few-shot learning (SF-CDFSL) problem. StepSPT first proposes a style prompt to adjust the target data distribution. It then employs a dual-phase optimization process with external and internal phases to optimize the style prompt. In the external phase, StepSPT proposes a step-wise distribution alignment while keeping the source model and classifier frozen. In the internal phase, the classifier is fine-tuned using cross-entropy loss, with the style prompt and source model fixed. Experimental results on five datasets show significant improvements by StepSPT over large pretrained models. Ablation studies validate the effectiveness of the style prompt and step-wise alignment.

Although StepSPT enables SF-CDFSL without fine-tuning LMs, it lacks strict constraints on feature certainty, particularly on EuroSAT and ISIC. This limitation is mainly due to the effects of resolution and background. Future work will focus on mitigating these challenges by addressing the impact of uncertain predictions on accuracy. For instance, we aim to explore methods that handle tasks across varying resolutions and assign greater weight to high-uncertainty samples (difficult samples) in the loss function, guiding models to learn optimal decision boundaries.

 
 \bibliographystyle{IEEEtran}
\bibliography{IEEEabrv,./amain}

\vspace{-35pt}

\begin{IEEEbiography}[{\includegraphics[width=1in,height=1.25in,clip,keepaspectratio]{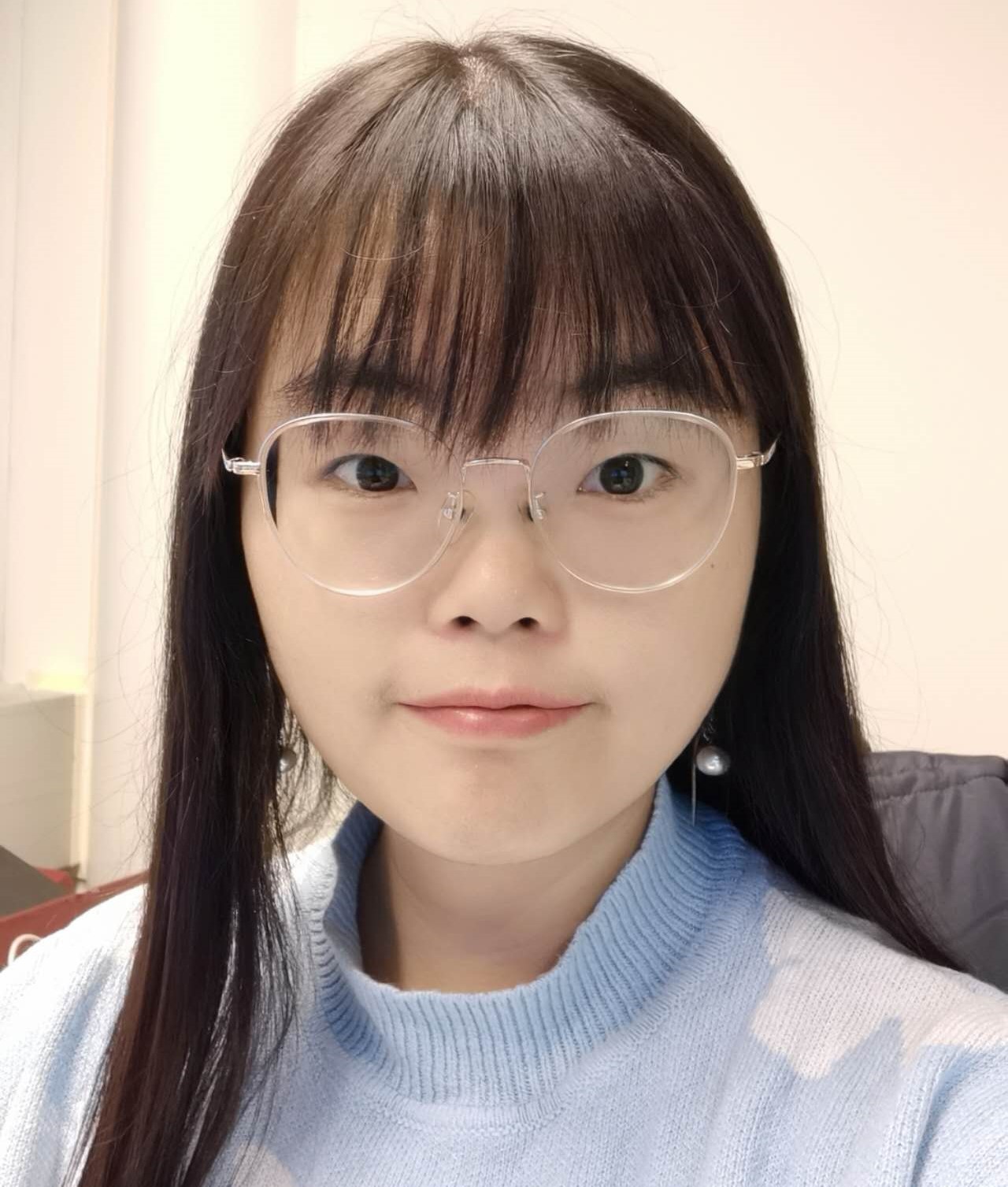}}]{Huali Xu}
 is currently pursuing the Ph.D. degree in Center for Machine Vision and Signal Analysis (CMVS) with the Faculty of Information Technology and Electrical Engineering, University of Oulu, Oulu, Finland. Her current research interests include computer vision, deep learning, few-shot learning, and cross-domain few-shot learning.
\end{IEEEbiography}

\vspace{-35pt}

\begin{IEEEbiography}[{\includegraphics[width=1in,height=1.25in,clip,keepaspectratio]{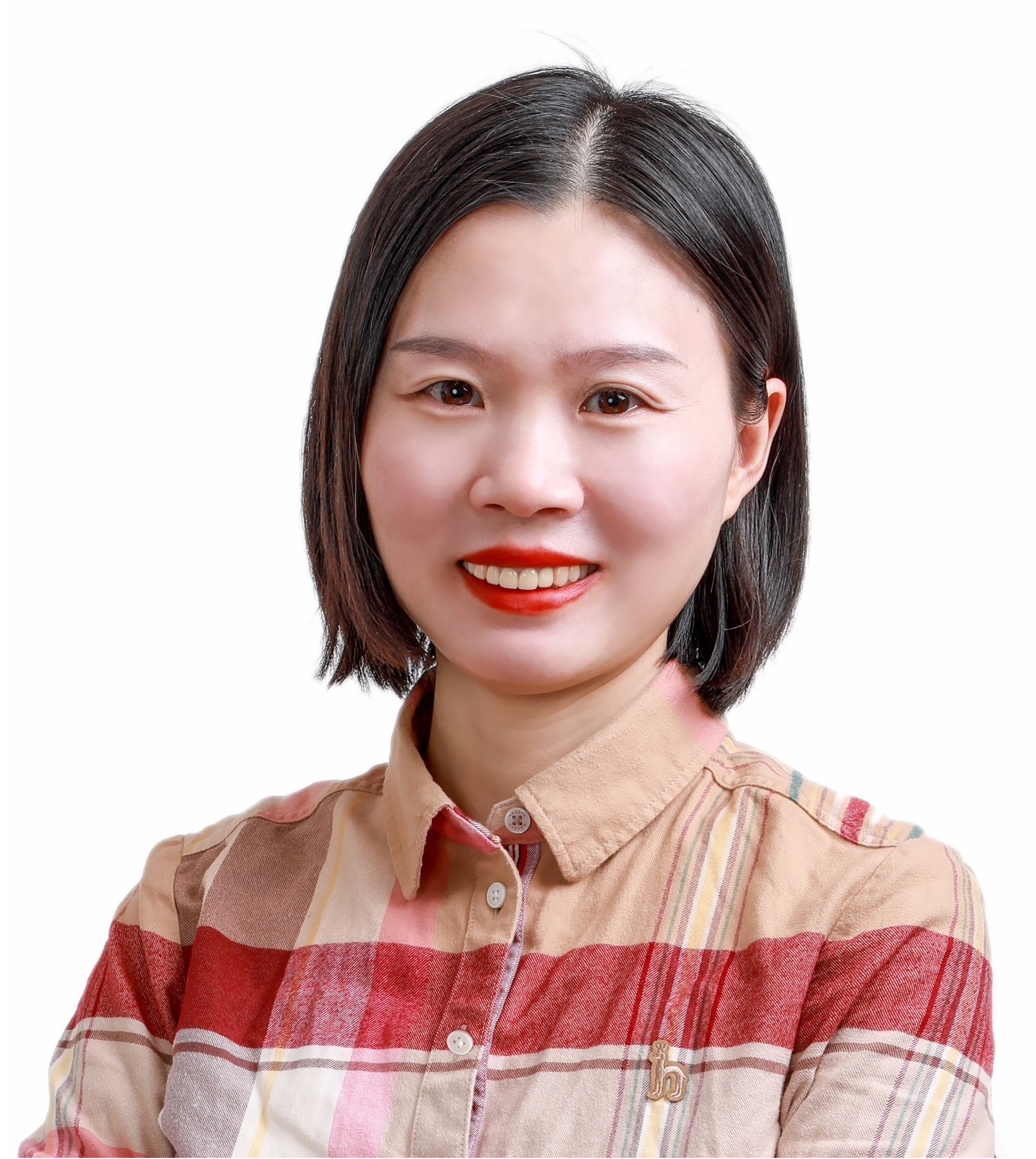}}]{Li Liu}
received the PhD degree in information and communication engineering from the National University of Defense Technology (NUDT), China, in 2012. She is now a full professor with the College of Electronic Science and Technology, NUDT. During her PhD study, she spent more than two years as a visiting student with the University of Waterloo, Canada, from 2008 to 2010. From 2015 to 2016, she spent ten months visiting the Multimedia Laboratory with the Chinese University of Hong Kong. From 2016 to 2018, she worked as a senior researcher with the Machine Vision Group, University of Oulu, Finland. She was a cochair of nine International Workshops with CVPR, ICCV, and ECCV. She served as the leading guest editor for special issues in IEEE TPAMI and IJCV. She is serving as the leading guest editor for IEEE PAMI special issue on “Learning with Fewer Labels in Computer Vision”. Her current research interests include computer vision, pattern recognition and machine learning. Her papers have currently more than 13000 citations according to Google Scholar. She currently serves as associate editor for IEEE TGRS, IEEE TCSVT, and PR.
\end{IEEEbiography}

\vspace{-33pt}

\begin{IEEEbiography}[{\includegraphics[width=1in,height=1.25in,clip,keepaspectratio]{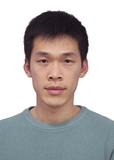}}]{Tianpeng Liu}
received the B.S., M.S., and Ph.D. degrees from the National University of Defense Technology (NUDT), Changsha, China, in 2008, 2011, and 2016, respectively. He is currently an Associate Professor with the College of Electronic Science and Technology. He has published numerous papers in respected journals, including IEEE Transactions on Aerospace and Electronic Systems and International Conference on Signal Processing. His primary research interests are radar signal processing, electronic countermeasure, and cross-eye jamming.
\end{IEEEbiography}

\vspace{-25pt}

\begin{IEEEbiography}[{\includegraphics[width=1in,height=1.25in,clip,keepaspectratio]{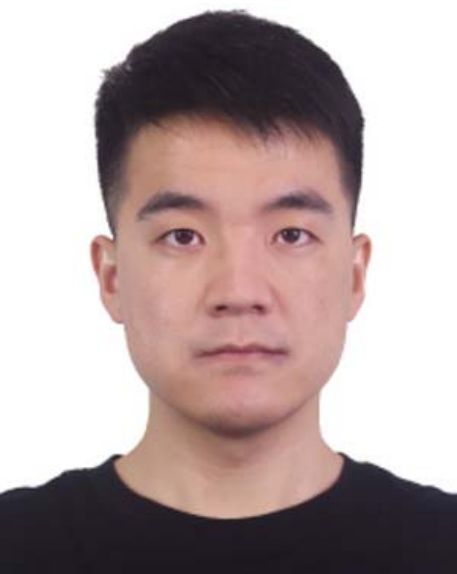}}]{Shuaifeng Zhi}
received his Ph.D. degree in Computing Research at the Dyson Robotics Laboratory, Imperial College London, UK, in 2021. He is currently a Lecturer (Assistant Professor) at the Department of Electronic Science and Technology, National University of Defense Technology (NUDT),
Changsha, China. He was a 6-month visiting student in 5GIC, University of Surrey, UK, in 2015. His current research interests focus on robot vision, particularly on scene understanding, neural scene representation, and semantic SLAM.
\end{IEEEbiography}

\vspace{-25pt}

\begin{IEEEbiography}[{\includegraphics[width=1in,height=1.25in,clip,keepaspectratio]{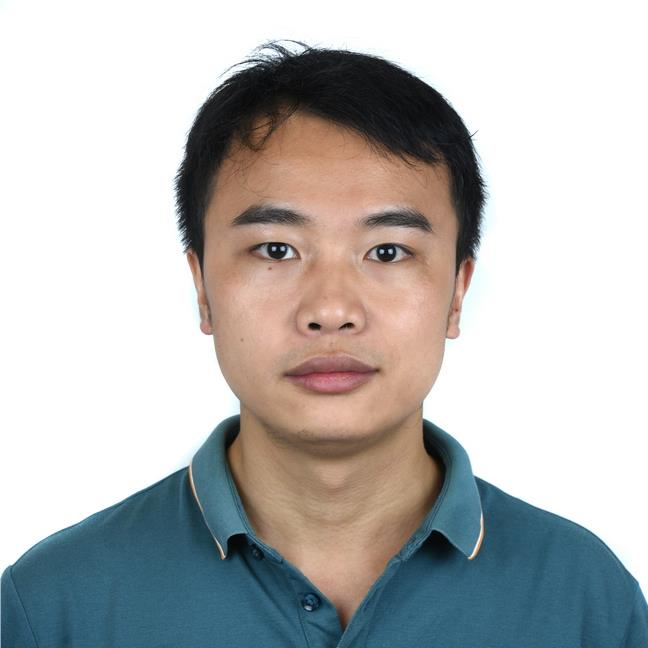}}]{Shuzhou Sun}
is currently working toward the PhD degree in center for machine vision and signal analysis (CMVS) with the Faculty of Information Technology and Electrical Engineering, University of Oulu, Oulu, Finland. His current research interests include computer vision, deep learning, and causal inference.
\end{IEEEbiography}

\vspace{-20pt}

\begin{IEEEbiography}[{\includegraphics[width=1in,height=1.25in,clip,keepaspectratio]{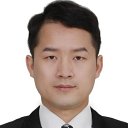}}]{Ming-Ming Cheng}
received his PhD degree from Tsinghua University in 2012. Then he did two years research fellow with Prof. Philip Torr in Oxford. He is now a professor at Nankai University, leading the Media Computing Lab. His research interests include computer graphics, computer vision, and image processing. He received research awards, including ACM China Rising Star Award, IBM Global SUR Award, and CCF-Intel Young Faculty Researcher Program. He is on the editorial boards of IEEE TPAMI/TIP.
\end{IEEEbiography}

\vfill

\end{document}